\pdfoutput=1

\documentclass[11pt]{article}
\usepackage[table,xcdraw]{xcolor}
\usepackage{acl}

\usepackage{subcaption}

\usepackage{xspace}
\usepackage{booktabs}
\usepackage{graphicx}
\usepackage{multirow}
\usepackage{tikz}
\usetikzlibrary{plotmarks}
\usepackage{fontawesome5}
\usepackage{tabularx}

\usepackage[normalem]{ulem}
\usepackage{times}
\usepackage{latexsym}
\usepackage{amsmath}
\usepackage{amsfonts}
\usepackage{tabularx}
\usepackage{cleveref}
\crefformat{section}{\S#2#1#3} 
\crefformat{subsection}{\S#2#1#3}
\crefformat{subsubsection}{\S#2#1#3}

\usepackage[T1]{fontenc}

\usepackage[utf8]{inputenc}

\usepackage{microtype}

\usepackage{inconsolata}
\usepackage{algorithm}
\usepackage{algorithmic}
\usepackage{soul}
\usepackage{enumitem}
\usepackage{todonotes}

\definecolor{darkgreen}{rgb}{0.0, 0.5, 0.0}
\definecolor{carnelian}{rgb}{0.7, 0.11, 0.11}

\usepackage{listings}
\newcommand{\DatasetName}{\textsc{MacGyver}}

\title{\textit{MacGyver}: Are Large Language Models Creative Problem Solvers?}

 \author{Yufei Tian$^{1}$\thanks{\ Work was done during Yufei's internship at AI2. Code and data available at: \url{https://github.com/allenai/MacGyver}} \quad  Abhilasha Ravichander$^{2}$ \quad Lianhui Qin$^{24}$ \quad Ronan Le Bras$^{2}$ \quad 
Raja Marjieh$^{3}$\\ 
 \textbf{Nanyun Peng}$^{1}$ \quad \textbf{Yejin Choi}$^{25}$ \quad \textbf{Thomas L. Griffiths}$^{3}$ \quad \textbf{Faeze Brahman$^{25}$}\\[7pt]
         $^1$University of California, Los Angeles, $^2$Allen Institute for Artificial Intelligence\\ $^3$Princeton University, $^4$University of California, San Diego,
         $^5$University of Washington\\[3pt]
         {\url{https://github.com/allenai/MacGyver}} \\[5pt]
         {
         \texttt{yufeit@cs.ucla.edu} \quad \texttt{faezeb@allenai.org}
         }
         }

\begin{document}
\maketitle
\begin{abstract}
We explore the creative problem-solving capabilities of modern LLMs in a novel constrained setting.
To this end, we create \DatasetName{}, an automatically generated 
dataset consisting of over 1,600 real-world 
problems deliberately 
designed to trigger \textit{innovative usage of objects} and necessitate \textit{out-of-the-box thinking}. We then present our collection to both LLMs and humans to compare and contrast their problem-solving abilities. \DatasetName{} is challenging for both groups, but in unique and complementary ways. For instance, humans excel in tasks they are familiar with but struggle with domain-specific knowledge, leading to a higher variance. In contrast, LLMs, exposed to a variety of specialized knowledge, attempt broader problems but fail by proposing physically-infeasible actions.
Finally, we provide a detailed error analysis of LLMs, and demonstrate the potential of enhancing their problem-solving ability with novel prompting techniques such as iterative step-wise reflection and divergent-convergent thinking. 

This work \textbf{(1)} introduces a fresh arena for intelligent agents focusing on intricate aspects of physical reasoning, planning, and unconventional thinking, which supplements the existing spectrum of machine intelligence; and \textbf{(2)} provides insight into the constrained problem-solving capabilities of both humans and AI. 

\end{abstract}

\section{Introduction}
\begin{figure}[t!]
    \centering
    \includegraphics[width=0.9\linewidth]{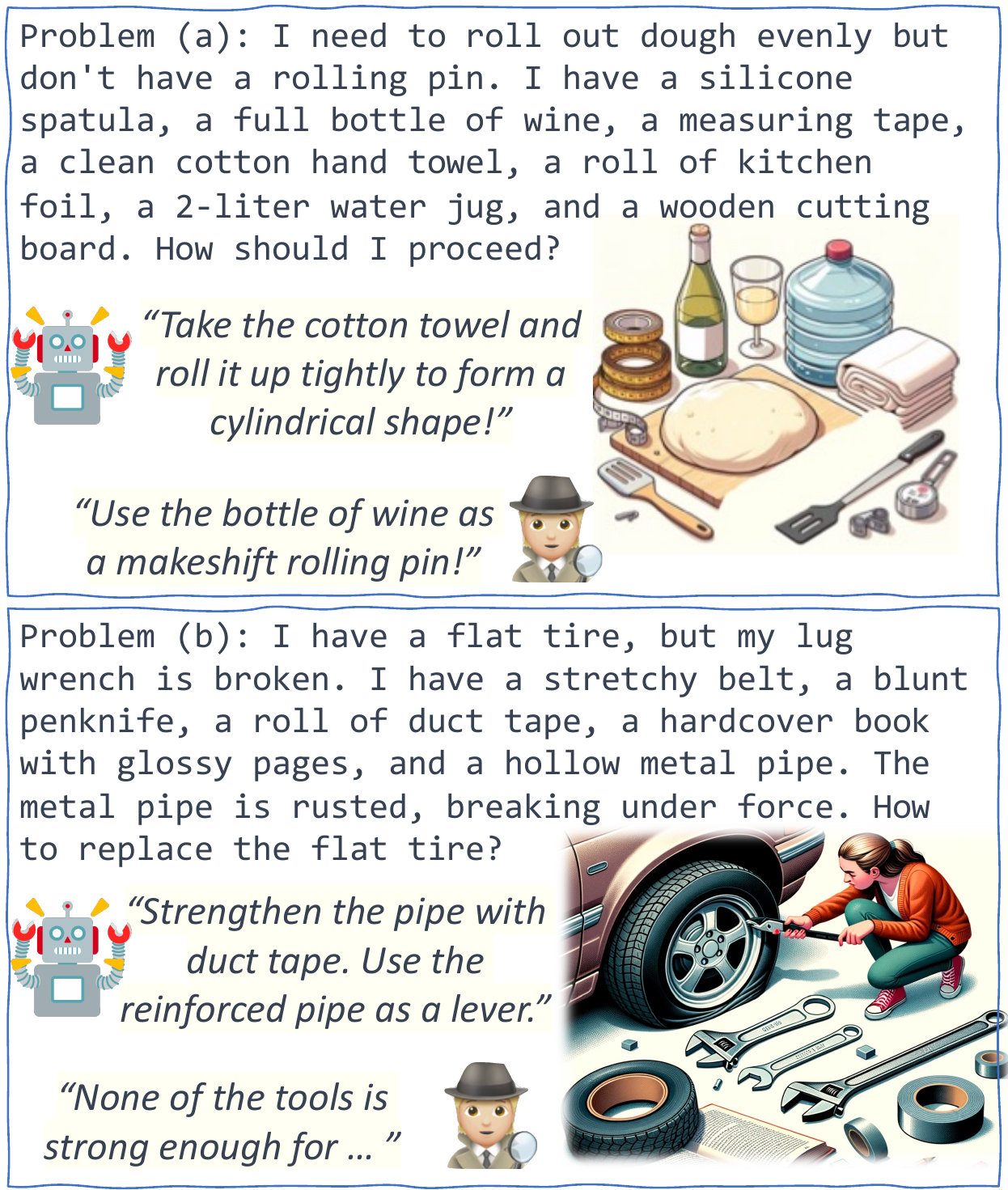}
    \vspace{-2mm}
    \caption{Examples of the problems in our \DatasetName{} dataset with the GPT-4 and human answers (continued in Figure \ref{fig:teaser_more}). 
    Pictures, drawn by DALL$\cdot$E 3, are solely for illustration purposes and may not accurately reflect the text. 
    In our experiment, all inputs to human and LLMs are natural language texts. 
}
    \label{fig:illustration}
    \vspace{-5mm}
\end{figure}

Creativity has long been considered the driving force behind modern civilization, and one of the hallmarks of human intelligence \cite{guilford1967nature,hennessey1995social}. As large language models (LLMs) have become increasingly powerful, 
researchers have begun to investigate their reasoning ability in problem-solving tasks \cite{yao2022react, brahman2023plasma} and 
their capacity for creativity as demonstrated by expressing humor and generating artistic content \cite{mittal-etal-2022-ambipun, hessel-etal-2023-androids, ramesh2022hierarchical, chakrabarty-etal-2022-help, tian-etal-2023-unsupervised}. 
However, everyday activities that involve creative thinking have not been studied to the same extent. 
In this work, we contribute a benchmark for creative problem solving, hoping to critically assess modern LLMs 
when it comes to `thinking out-of-the-box'.


To bridge this gap, we curate \DatasetName{}, \textbf{\textit{a novel unconventional problem-solving dataset}} consisting of 1,683 sets of verbal problems that require human-like creativity in the realm of physical reasoning.
Drawing inspiration from the cognitive science literature \cite{duncker1945problem}, we collect problem scenarios that deliberately push against \textit{functional fixedness}\textemdash a cognitive bias 
that limits an agent from employing familiar tools in innovative ways. Notably, leveraging the \textit{generative} strength of LLMs and the \textit{verification} strength of humans, we design a novel and labor-efficient pipeline to collect progressively more challenging scenarios (\cref{sec:data_collect}). These scenarios are verified by humans as requiring unconventional usage of objects to find a solution. For example, solving problem (a) in Figure \ref{fig:illustration} requires using the wine bottle as a makeshift rolling pin.\footnote{If the problem is unsolvable given the presented tools and constraints (problem b in Figure \ref{fig:illustration}), we expect the agent to identify such infeasibility and provide a short justification.}  Each problem in our dataset is paired with at least one human-provided or verified solution. To the best of our knowledge, \DatasetName{} is the first dataset of unconventional everyday problems requiring two key elements of creativity \cite{guilford1967creativity}: \textit{\textbf{divergent}} thinking (to come up with creative or unconventional usage of objects) and \textit{\textbf{convergent}} thinking (to accomplish a goal efficiently).

Next, we use the resulting dataset as a \textbf{\textit{benchmark}} to evaluate the creative problem-solving abilities of both human participants and recent LLMs, including GPT-3.5, GPT-4, PaLM2, Claude2, and Llama2 \cite{chatgpt2022,openai2023gpt4,anil2023palm, touvron2023llama,claude2}. 
Our results in \cref{sec:benchmark} reveal a substantial gap between most LMs and human. While the best performing LM, GPT-4, complements the capability of an arbitrary human under certain domain-specific settings (\textit{e.g., fixing a hole on the wall}), humans' collective wisdom is so far still invincible. 
Additionally, LLMs struggle to identify unsolvable problems and either exhibit misleading helpfulness 
or are ultraconservative in inappropriate cases. In \cref{sec:compare_human_machine}, we present detailed comparison between human and machine.

Finally, a qualitative analysis of LLM responses reveals two common \textbf{\textit{failure modes}}: (1) models propose physically infeasible, unnecessary, or wrong solution steps that deviate from the intended goal, and 2) models hallucinate unavailable tools or do not adhere to constraints specified. We propose two \textbf{\textit{prompting strategies}} to mitigate these common error types: (1) a self-reflection based strategy to iteratively verify the feasibility of each generated step and then modify as necessary, and 2) a cognitive-science-inspired strategy of first divergently exploring the potential use of presented tools and then converging on the problem solution. Experimental results show the efficacy of both strategies in boosting models performance (\cref{sec:enhance}). We hope \DatasetName{} will serve as a useful resource for
\begin{itemize}[leftmargin=0pt]
    \item Evaluating LLMs and autonomous agents in new challenges involving real-world scenarios, innovative object usage, and physically feasible actions;
    \item Enhancing LLMs’ creativity and physical-related reasoning skills; and
    \item Providing useful insight and resources to researchers in other fields such as computational cognition and psychology
\end{itemize}


\section{\DatasetName{} Dataset}\label{sec:data_collect}
\begin{figure*}[ht!]
    \centering
    \includegraphics[width=0.97\linewidth]{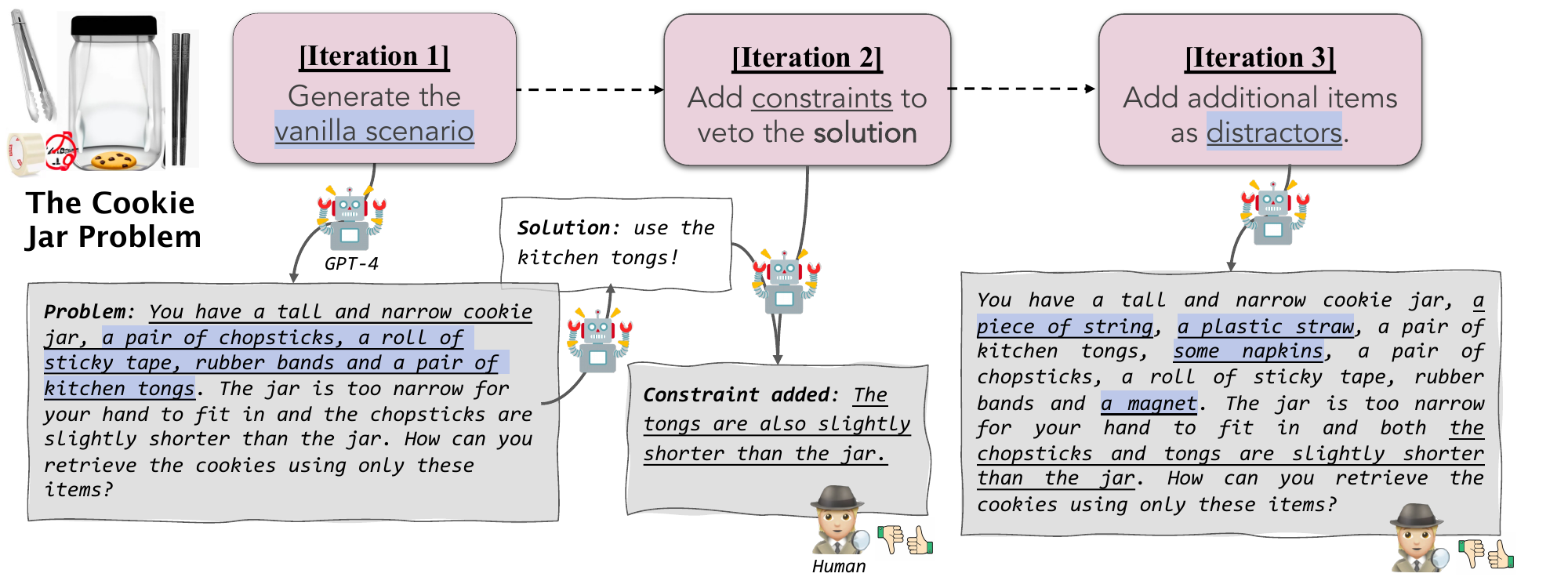}
    \vspace{-3mm}
    \caption{Progressive problem refinement with GPT-4. Starting from a vanilla version (\textit{i.e.,} Iteration 1), we carefully design refinement steps that gradually increase the problem's complexity by adding specific object properties as constraints to veto a previous solution (\textit{i.e.,} Iteration 2), and adding distracting objects that are (likely) not involved in the solution the problem (\textit{i.e.,} Iteration 3). After that, human verifiers judge the quality of refined problems. }
    \vspace{-3mm}
    \label{fig:progressive_data_creation}
\end{figure*}

LLMs have demonstrated utility for idea generation \cite{girotra2023ideas}. Therefore, instead of asking humans to come up with thousands of constrained scenarios from scratch, we design a progressive refinement pipeline to explore LLMs' potential to generate problem settings quickly and at scale (\S \ref{subsec:progressive_data_creation}). Human annotators then verify that each problem is concrete and requires creativity (\cref{subsec:human verification}). Each instance in our dataset includes a constrained problem setting
paired with at least one human-provided or verified solution (\cref{subsec:human verification}, \S \ref{subsec:collect_gold}). 

\subsection{Progressive Problem Refinement for Dataset Creation 
}\label{subsec:progressive_data_creation}

Figure \ref{fig:progressive_data_creation} provides an illustration of our problem collection pipeline, showing how we combine human and machine inputs. Specifically, we propose a progressive problem refinement approach that gradually increases problem complexity by 1) adding specific object properties (\textit{e.g.,} material, size, etc.) as constraints to eliminate a previous solution and 2) adding distracting objects that are not involved in the solution. From a cognitive perspective on problem-solving \cite{knoblock1991search}, the first refinement step removes the most straightforward solution path, while the second step further complicates the problem by adding branches to the search space.

We implement this pipeline through a dialogue interaction with GPT-4. Human assessment results (detailed in \cref{subsec:result_3_iterations}) confirm that both steps within the progressive refinement approach pose additional challenges to LLMs, and after the two iterations, the original problem requires more creativity and becomes more challenging. 

\subsection{Human Verification Process}\label{subsec:human verification}

After the refinement process, we involve human verifiers to judge if the final versions of the problems \textbf{1)} are solvable, unsolvable, or need more clarification (\textit{e.g.,} the setup is vague, which will be discarded), and \textbf{2)} for those solvable, whether solving them efficiently requires creative thinking (\textit{i.e.,} using objects to achieve goals they were \textit{not} originally designed for \textemdash unconventional usage). Each problem is annotated by three human verifiers, with average inter-annotator agreement (IAA, measured by Cohen's Kappa) of 0.67 and 0.77 for tasks \textbf{1)} and \textbf{2)}, respectively. 
Finally, we pair each problem with a gold answer. For the solvable subset, it is a step-by-step feasible solution. For the unsolvable subset, it is an explanation why the stated goal cannot be achieved (detailed in \S \ref{subsec:collect_gold}).
\begin{table}[]
\centering
\small

\begin{tabular}{@{}lccc@{}}
\toprule
\textbf{Problem (All)}   & \textbf{Solvable} & \textbf{Unsolvable} & \textbf{Total} \\ \midrule
Count      & 1,306      & 377        & 1,683  \\
Percentage & 77.6\%   & 22.4\%     & 100\% \\ \bottomrule
\end{tabular} \\

\begin{tabular}{@{}lccc@{}}
\toprule
\begin{tabular}[l]{@{}l@{}}\textbf{Problem} \\ \textbf{(Solvable  Subset)}\end{tabular} & \textbf{Unconv.} & \textbf{Conv.} & \textbf{Total}   \\ \midrule
Count                                                                 &1,073            & 233          & 1,306     \\
Percentage                                                            & 82.2\%         & 17.8\%       & 100.0\% \\ \bottomrule
\end{tabular}
\vspace{-2mm}
\caption{Statistics of the entire \DatasetName{} dataset (top). Number of solvable problems that require unconventional use of tools (bottom).}
\vspace{-5mm}
\label{table:stats_of_data}

\end{table}

 In total, we created 1,683 problems, with a detailed breakdown in Table \ref{table:stats_of_data}. Of those, 78\% are solvable and 22\% are unsolvable. Another 7\% of all problems were discarded after being annotated by at least one annotator to be ambiguous or contradictory. For solvable problems, 82\% require using tools in an innovative or unconventional manner.

\subsection{Diversity Control and Check} \label{subsec:diversity_control}

Intuitively, we want to avoid generating multiple problems with familiar goals and constraints. 
In this section, we summarize our measures to ensure the collected problems are \textit{diverse, comprehensive, and free of repetitive patterns}.

\paragraph{Diversity Control} We hand-craft more than 50 tags of locations and activities, aiming to ensure that our data collection pipeline delves into a variety of topics. 
These predefined tags are integrated into the prompt that we used to query GPT-4 for problem curation at Iteration 1
. The detailed list of all tags can be found in  Table \ref{table:tags}. 

\begin{figure}[t!]
\centering
    \hspace*{-3mm}\includegraphics[width=0.5\textwidth]{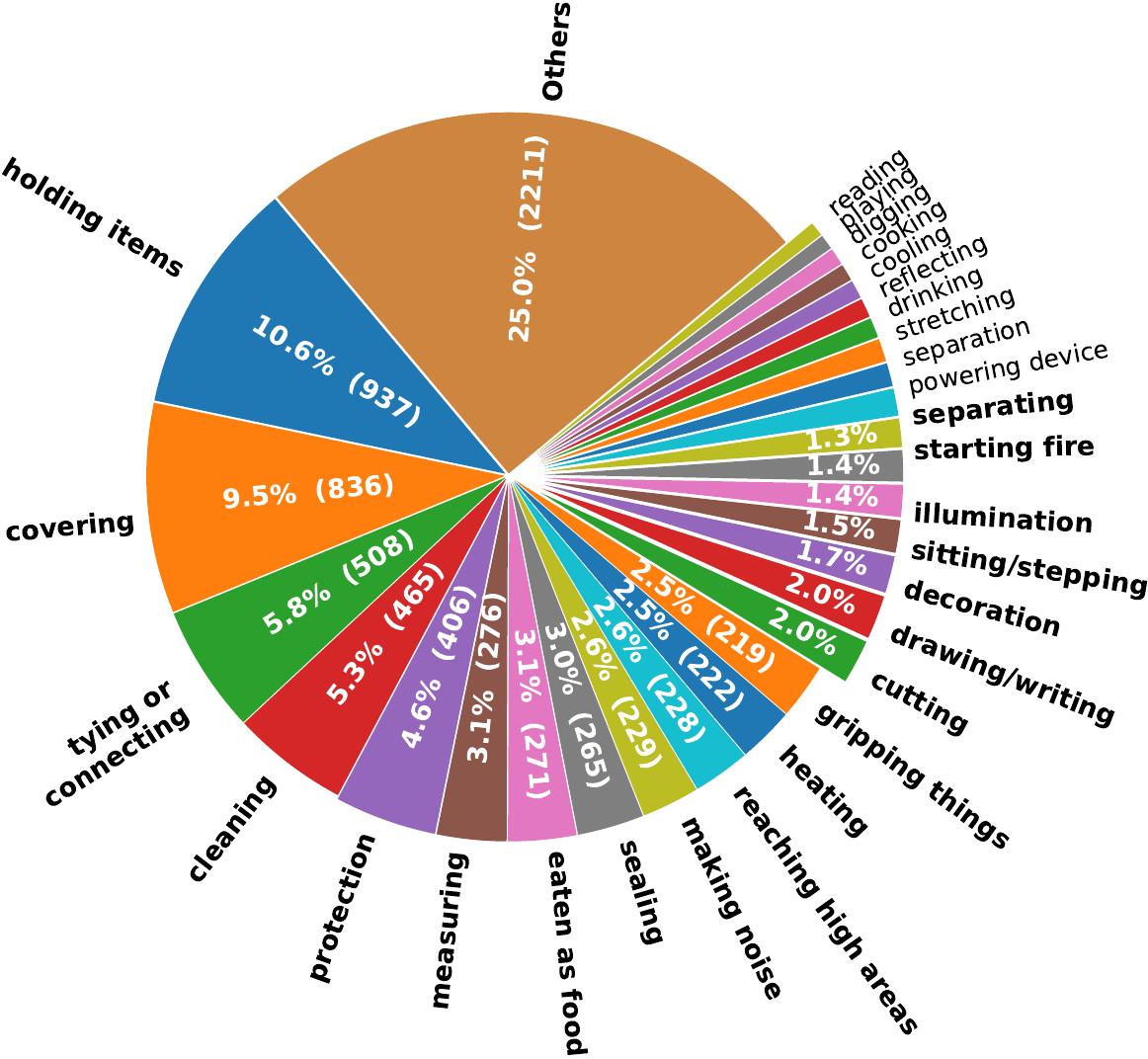}
    \vspace{-7mm}
    \caption{Affordances of the presented tools in our \DatasetName{} dataset and their frequency (and count). Note that one object may have multiple affordances (\textit{e.g.}, paddle boards can be used for boating, reaching high areas, and exercise).
    }
    \label{pie-chart:affordance}
    \vspace{-6mm}
\end{figure}

\paragraph{Diversity Check} After the final iteration, we parse the objects presented as tools among all generated problems. Intuitively, we consider two similar objects with different properties (\textit{e.g.,} \textit{plastic knife} and \textit{metal knife}; \textit{eyeglasses} and \textit{magnifying glass}) to be different. In total, 3,800 unique tools were identified. We compute their frequency and use GPT-4 to analyze their affordances (Appendix Table \ref{table:common_tools}; Figure \ref{pie-chart:affordance}). We found that \textit{holding items} and \textit{covering} are the top two types, followed by \textit{tying or connecting}  and  \textit{cleaning}. The long tails in both illustrations signify a desirable level of diversity.\footnote{Refer to Appendix \ref{appendix:diversity_control} for more details such as the detailed list of all tags, the most frequent tools and their affordances, and the prompt used to analyze tool affordance.}

\section{Assessing the Task Difficulty}\label{sec:assessment_setup}

To gauge the challenge of our task posed to the most recent LLMs, we evaluate the zero-shot performance of GPT-4 \cite{openai2023gpt4}. Nevertheless, existing automatic evaluations fall short to assess the efficacy of a presented solution. Therefore, we recruit human annotators to evaluate the quality of the GPT-4's answers on the \textbf{\textit{entire}} \DatasetName{}.

\paragraph{\textbf{Assessment Setup}. }For a solvable problem, human annotators are asked to judge if the presented solution is \textbf{1.1} \textit{feasible and efficient}\footnote{A solution is considered efficient if it has no redundant or unnecessary steps, and it is unlikely that the problem can be solved with less labor or using fewer steps.}, \textbf{1.2} \textit{feasible yet inefficient}, or \textbf{1.3} \textit{infeasible}. The machine-generated answer may also wrongly assume the problem is unsolvable and gives a wrong justification (\textbf{1.4}). For an unsolvable problem, they need to judge if the presented answer \textbf{2.1} \textit{correctly identifies the problem as unsolvable}, and \textbf{2.2} \textit{gives the right justification}. Similarly, the answer may also wrongly assume the problem is solvable and give a wrong solution (\textbf{2.3}).

\begin{figure}[t!]
    \centering
    \includegraphics[width=0.99\linewidth]{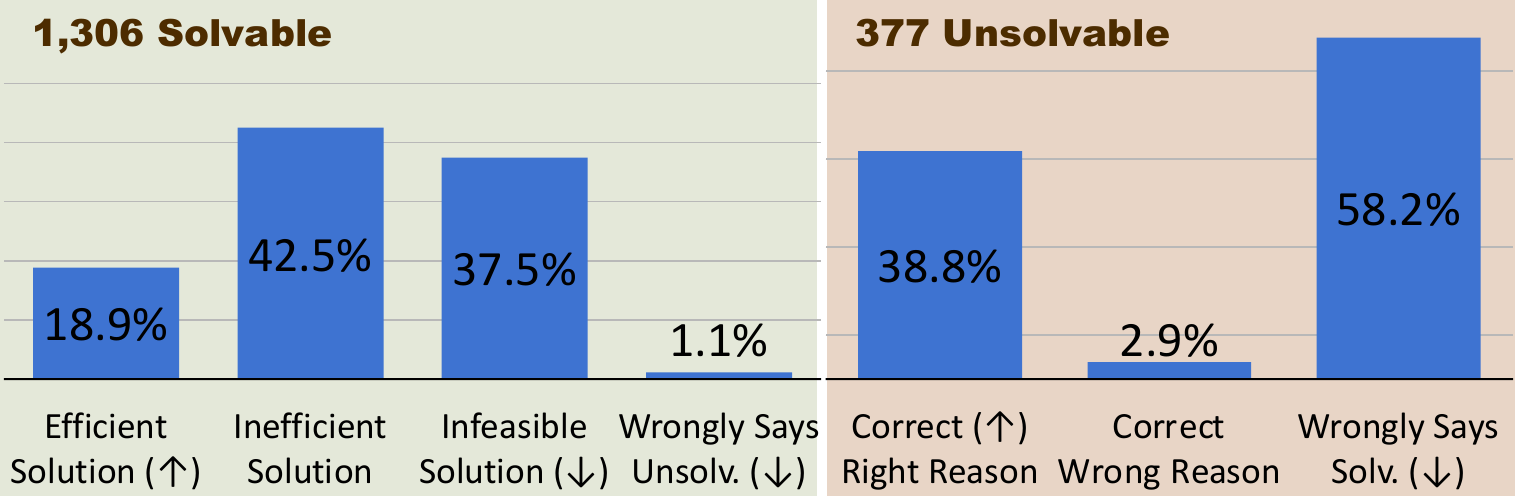}
    \vspace{-5mm}
    \caption{Left: Human-evaluated GPT-4 performance on all 1,306 problems from the \DatasetName{} that humans think are \textbf{solvable}. Right: GPT-4 performance on all 377 problems that humans think are \textbf{unsolvable}. \uline{\textit{Correct for the right reason}} means that the LLM correctly identifies the problem is unsolvable, and gives the right justification. \uline{\textit{Correct for the wrong reason}} means that it correctly identifies the problem is unsolvable, but gives an incorrect justification. 
}
    \label{fig:assess_difficulty}
    \vspace{-5mm}
\end{figure}

\paragraph{\textbf{GPT-4 Performance.} }We report the performance on the solvable and unsolvable subset in Figure \ref{fig:assess_difficulty}. Our preliminary findings indicate that, \textbf{firstly}, LLMs as strong as GPT-4 still exhibit limitations in solving unconventional problems, with only 18.9\% likelihood of providing an efficient solution, while 37.5\% likelihood of providing an infeasible solution. Analysis in the later section (\cref{sec:enhance}) shows that one common mistake is it failing to realize the consequences of actions and tool affordances in the given context (\textit{e.g.,} proposing to use chopsticks to lift up the egg yolk). \textbf{Secondly}, GPT-4 displays overconfidence, often suggesting solutions to problems that are inherently unsolvable. This could be partially due to GPT-4 being trained with RLHF \cite{NEURIPS2022_b1efde53}, maximizing its helpfulness. Moreover, the model struggles to discern whether a problem description is sufficiently concrete for resolution or too ambiguous, necessitating additional context \cite{liu-etal-2023-afraid}.

\section{Benchmarking Humans and LLMs}\label{sec:benchmark}
A natural follow-up question is how well modern LLMs perform on this task, as compared to humans. We thus evaluate the performance of several recent LLMs (\textit{i.e.,} PaLM2, Claude2, Llama2, GPT-3.5 and GPT-4) 
on a representative sample of the entire \DatasetName{} dataset which contains 323 problems. In addition, we gauge the capability of average humans on the same set of tasks.

\subsection{Collecting Independent Human Responses}\label{subsec:human_study_prolific}

We assessed human capability by recruiting participants who are new to this task. To this end, independent solutions were collected from a pool of $N=252$ UK participants on \href{https://www.prolific.co}{Prolific}. We intentionally used a different platform and target population from those of the human evaluators (\textit{i.e.}, MTurk and US) to minimize any chances of overlap. For a given problem, participants indicated whether they believed the problem is solvable, unsolvable, or required further clarification. If solvable, they provided a step-by-step solution, and otherwise explained why the problem was unsolvable. Overall, we elicited an average of six responses per problem and each participant contribute to up to five different problems.

\subsection{Collecting Machine Responses}\label{subsec:collect_machine_sol}
\begin{figure*}[t!]
    \centering
    \includegraphics[width=0.95\linewidth]{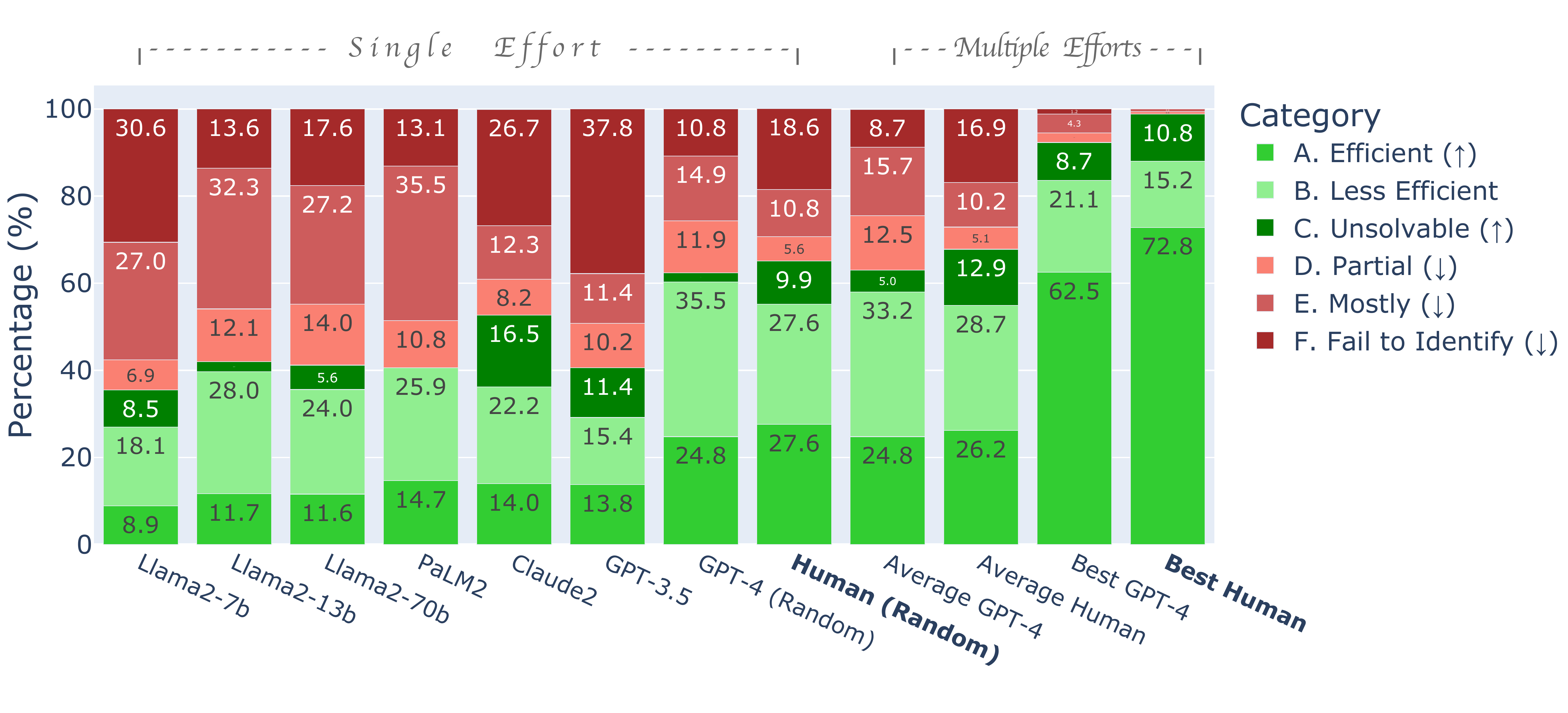}
    \vspace{-8mm}
    \caption{\textbf{Left}: Benchmark results of seven LLMs and human with a single effort.  For human participants, since there is no single participant who worked on all problems, we take a random response from each problem. We color-code the three categories indicating fine-grained aspects of \textcolor{darkgreen}{correctness} or \textcolor{carnelian}{falseness}. 
    \textbf{Right}: Comparison between GPT-4 and human where we evaluated multiple solutions per problem. The best performance, which can be viewed as an upper bound, is computed by taking the individual best answer (out of 6) for each problem. The actual numbers are reported in Table \ref{Table:benchmark-results} in \cref{appendix:benchmark}.
}
    \label{fig:benchmark-results}
    \vspace{-5mm}
\end{figure*}

We collected solutions from seven different LLMs using Nucleus sampling \cite{HoltzmanBDFC20} and return the top one sequence ($T$=$0.7$ and $p$=$0.95$).
In the prompt, we instruct an LLM to either provide a feasible and efficient solution to a problem when it believes the problem is solvable, or otherwise a justification explaining why the given problem is unsolvable.
To explore whether different sizes of the same model plays a role in its problem solving ability, we include three variations of Llama2 (\textit{i.e.}, \texttt{-7b}, \texttt{-13b}, \texttt{-70b}), as well as two variants of GPT model family (\textit{i.e.}, \texttt{gpt-3.5-turbo}, \texttt{gpt-4-0613}).

\paragraph{Additional GPT-4 Responses}
For a fair comparison with humans,
we emulate the same setup in \cref{subsec:human_study_prolific} by obtaining multiple solutions per problem from a single LLM. Since exhaustive human evaluation is costly, we opted to elicit multiple solutions exclusively from the most capable LLM, GPT-4. Multiple manually-designed instructions are
used to prompt GPT-4 in order to reduce repetition among separate sessions of API calls. More details can be found in Appendix \ref{appendix:experiment_setup}.

\subsection{Human Evaluation}\label{subsec:human_eval}
Human annotators were asked to evaluate if a presented answer is correct by selecting one out of six fine-grained categories: \textbf{A (or B)} correctly giving a feasible and efficient (or less efficient) solution to a solvable problem; \textbf{C} correctly identifying an unsolvable problem and giving the right justification; \textbf{D} giving a partially incorrect answer; \textbf{E} giving a mostly or entirely wrong answer; and \textbf{F} failing to identify the correct solvability status.
\footnote{Screenshots of  the human evaluation interface can be found in Appendix Figure \ref{fig:survey_benchmark_01} and \ref{fig:survey_benchmark_02}.}

\subsection{Benchmark Results}

We report the benchmark results in Figure \ref{fig:benchmark-results}. Category \textbf{A}, \textbf{B}, and \textbf{C} are the three aspects of correct responses, while the remaining \textbf{D}, \textbf{E}, and \textbf{F} are aspects of the wrong ones. At a glance, despite varying in their characteristics, all of the benchmarked LLMs lag behind the performance of humans.

\subsubsection{Performance with Single Effort} 
As is mentioned in \cref{subsec:collect_machine_sol}, only the top one response is collected for a LLM per problem. Hence, we first list the LLMs' performances with their \textit{\textbf{single best answers}} on left of Figure \ref{fig:benchmark-results}.
 For human participants, there is no single person who approached all problems. Therefore, to simulate \textit{\textbf{an arbitrary person's individual}} performance, we take a random response from each problem.

We see that most recent LLMs achieve a mere 35\% to 42\% chance of success. Although GPT-4 and Claude2 stand out among the tested LLMs, their best attempts still under-perform an arbitrary average person with total correct rate of 65.1\% (sum of category \textbf{A}, \textbf{B} and \textbf{C}). 

We observe that different families of LLMs exhibit dissimilar behaviors. For example, PaLM2 and GPT-4 are overly verbose and often suggest solutions to problems that are inherently unsolvable (as seen by their remarkably low performance in category \textbf{C}: correctly identify an unsolvable problem). In contrast, Llama2-7b, Claude2, and GPT-3.5 are more conservative and fail to realize a constrained problem can still be solvable (reflected in their high numbers in category \textbf{F}). Comparing the three variants of Llama2, we find that the larger models (13b, 70b) excel in correctly identifying solvability (category \textbf{F}). The smaller model (7b) is more subject to falsely recognizing a constrained problem as unsolvable. Beyond this, however, it appears that scale alone does not significantly unleash any creative problem-solving capabilities.

\subsubsection{Performance with Multiple Efforts}
Recall that we 
collect multiple solutions per problem for GPT-4 and humans. With these, we compute the \textit{\textbf{average}} and \textit{\textbf{best}} performance. The best performance, which can be viewed as an upper bound, is computed by taking the individual best answer for each problem. The results are shown on the right of the same figure. In addition, we compute the \textit{\textbf{majority}} performance by considering a binary annotation (\textit{i.e.,} correct or wrong) of each problem. We find that the majority of humans are 79.3\% correct, surpassing that of GPT-4 (73.3\%).

We see that on average human participants are slightly worse than GPT-4 in coming up with a correct solution (especially inefficient ones, category \textbf{B}), which is potentially owing to functional fixedness. In general, humans still out-perform GPT-4 due to the fact that GPT-4 seldom correctly identifies an unsolvable problem. 
Moreover, the best of the four human answers, which can be considered as a form of \textit{collective wisdom}, clearly leads to a near perfect performance.

Finally, humans seem to struggle with certain problems (category \textbf{F}). We hypothesize that an individual person, who likely does not have domain-specific knowledge in all aspects of life, may not outperform a single LLM such as GPT-4, which is trained on massive amount of data and a wide variety of tasks. However, 
when considered collectively as a group, with each person contributing their unique expertise and wisdom, human intelligence exceeds that of LLMs. To verify our hypothesis and gain deeper insights into 
the relationship between the intelligence of humans and LLMs, we conduct further analyses in the next section.

\section{Comparing GPT-4 with Humans}\label{sec:compare_human_machine}
\subsection{Humans have higher variance than LLMs.}
\begin{figure}[t]
\centering
    \includegraphics[width=0.45\textwidth]{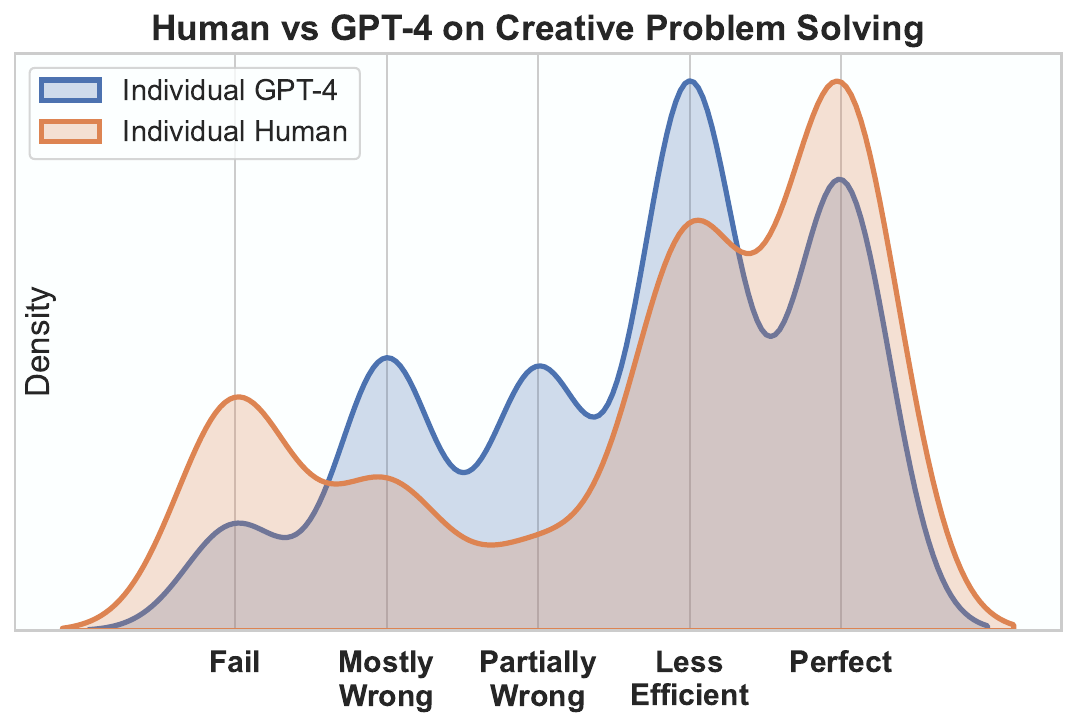}
    \vspace{-3mm}
    \caption{The kernel density estimate of individual human and GPT-4 answers.
    }
    \vspace{-3mm}
    \label{figure:compare_variance}
\end{figure}

We plot the kernel density estimate (KDE) of individual human and GPT-4 responses in Figure \ref{figure:compare_variance}. We can see that humans either approach a problem perfectly or fail totally. Namely, once humans understand the task and acquire the relevant knowledge, they can always propose a feasible and often the most efficient solution. 
On the contrary, GPT-4 responses fall more into the middle (mostly/partially wrong, or inefficient), owning to its ability to aggregate information from a wide range of sources it has been trained on. However, GPT-4 is sometimes ignorant of tool affordances or consequences of its proposed actions, lacking the depth of understanding that humans possess (see more detailed error analysis in \S\ref{subsec:error_analysis}).


\subsection{Humans possess better general everyday knowledge, but less domain-specifically.}\label{subsec:compare_2d_plot}

\begin{figure}[t]
    \centering
    \includegraphics[width=1.0\linewidth]{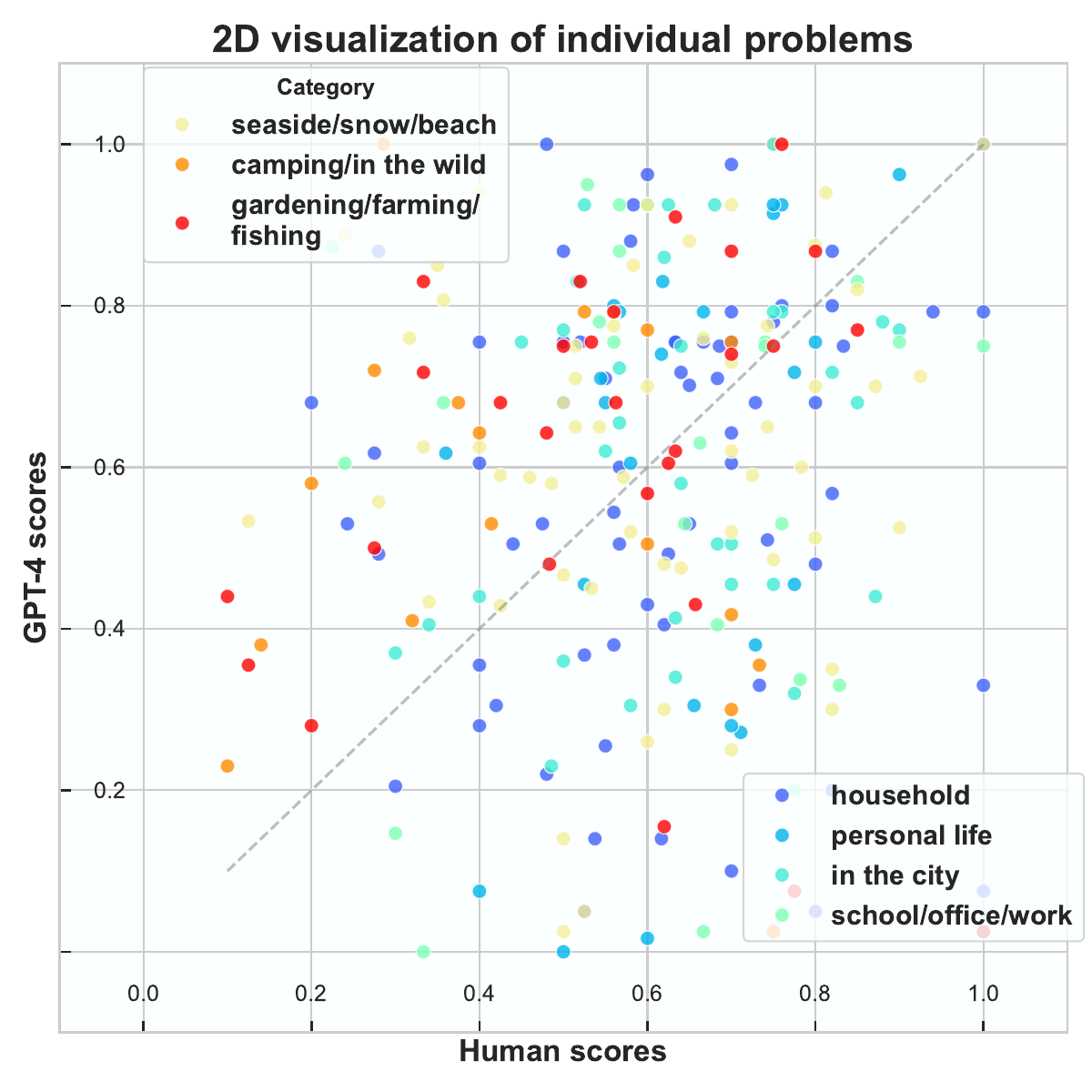}
    \vspace{-8mm}
    \caption{2D visualization of human (x-axis) and GPT-4 (y-axis) performance on individual problems. Each dot represents a problem, with its color representing seven different categories. Humans are better at solving problems that they are familiar with (\textit{e.g.,} household), than those requiring domain-specific knowledge (\textit{e.g.}, gardening/farming/fishing). 
}
\vspace{-5mm}
    \label{fig:problem_comparison}
\end{figure}

Next, we visualize the capability of humans and GPT-4 on individual problems in a 2D plot (Figure \ref{fig:problem_comparison}). Accordingly, we convert categorical labels into numerical scores ranging from 0 (Fail) to 1 (Perfect), and take the average score across solutions.  
We also plot the diagonal line: the farther away a point is from this, the larger the gap between human and GPT-4 performance.
\begin{table*}[t!]
\centering
\renewcommand{\arraystretch}{1.4}
\footnotesize
\begin{tabularx}{\linewidth}{>{\hsize=1.1\hsize}X>{\hsize=0.9\hsize}X c}
\toprule
\textbf{Error Description}                                          & \textbf{Example}                                        & \textbf{Freq.} \\ \midrule
\rowcolor[HTML]{E7E6E6}
  \textbf{(1) Wrong tool usage.}  Using tools in ways that are physically infeasible or not afforded &
  Using the stapler to staple the duct tape on top of broken glasses. & 42.4\% \\

  \textbf{(2) Not achieving the goal.} The proposed approach contains unnecessary or wrong steps towards the stated goal  &
  To save space when packing, use the scissors to cut the comforter into smaller pieces. & 17.7\% \\
\rowcolor[HTML]{E7E6E6} 
 \textbf{(3) Using unavailable tools.} 
 & - & 16.9\%  \\
\textbf{(4) Wrong spatial understanding}  & Putting the shoe box inside the empty DVD case. & 10.8\%  \\
\rowcolor[HTML]{E7E6E6}
\textbf{(5) Unfaithful to constraints. }Ignoring constraints added to a tool or a situation  & -  & 9.5\%  \\ \bottomrule
\end{tabularx}
\vspace{-3mm}
\caption{Categories of common errors made by GPT-4. It is highly prone to coming up with actions that are physically infeasible, unnecessary, or wrong. An erroneous solution may have more than one type of mistake. }
\vspace{-3mm}
\label{table:error_analysis}
\end{table*}

We find that humans are better at solving tasks in categories likely to be familiar to them, such as \textit{household} and \textit{personal life}. For those requiring domain-specific knowledge such as \textit{gardening/farming/fishing}, GPT-4 performs better. The same 
holds when we manually inspect the outliers: those few problems that belongs to everyday categories yet humans are poor at. 
Unsurprisingly, they are problems such as demonstrating the concept of refraction without a prism (category: school),  and making a sundial (category: beach), which an average person might have little experience with. Refer to \S \ref{appendix:subsec:compare} for examples and other comparisons.

Overall, the different creative strengths of humans and AI systems suggests that the most effective solutions to tasks requiring thinking ``out-of-the-box'' might arise from a \textit{collaborative approach} leveraging the strengths of both parties.

\section{Enhancing LLMs' Problem Solving}\label{sec:enhance}
Here, we investigate whether different prompting strategies
can enhance the problem-solving abilities of existing LLMs. In \cref{subsec:error_analysis}, we conduct a detailed error analysis on GPT-4, showing it is weakest at identifying the correct tool affordance and physical feasibility. In \cref{subsec:improve_via_prompting}, we propose two new prompting strategies that effectively reduce its mistakes.
\subsection{Error Analysis for GPT-4}\label{subsec:error_analysis}

To better understand the limitations of LLMs and provide insight for potential improvement, we 
manually analyze
200 solutions generated by GPT-4 marked as infeasible by human annotators. We identified five common failure modes in Table \ref{table:error_analysis}. 

\begin{figure*}[t!]
\vspace{-2mm}
    \centering
    \includegraphics[width=0.96\linewidth]{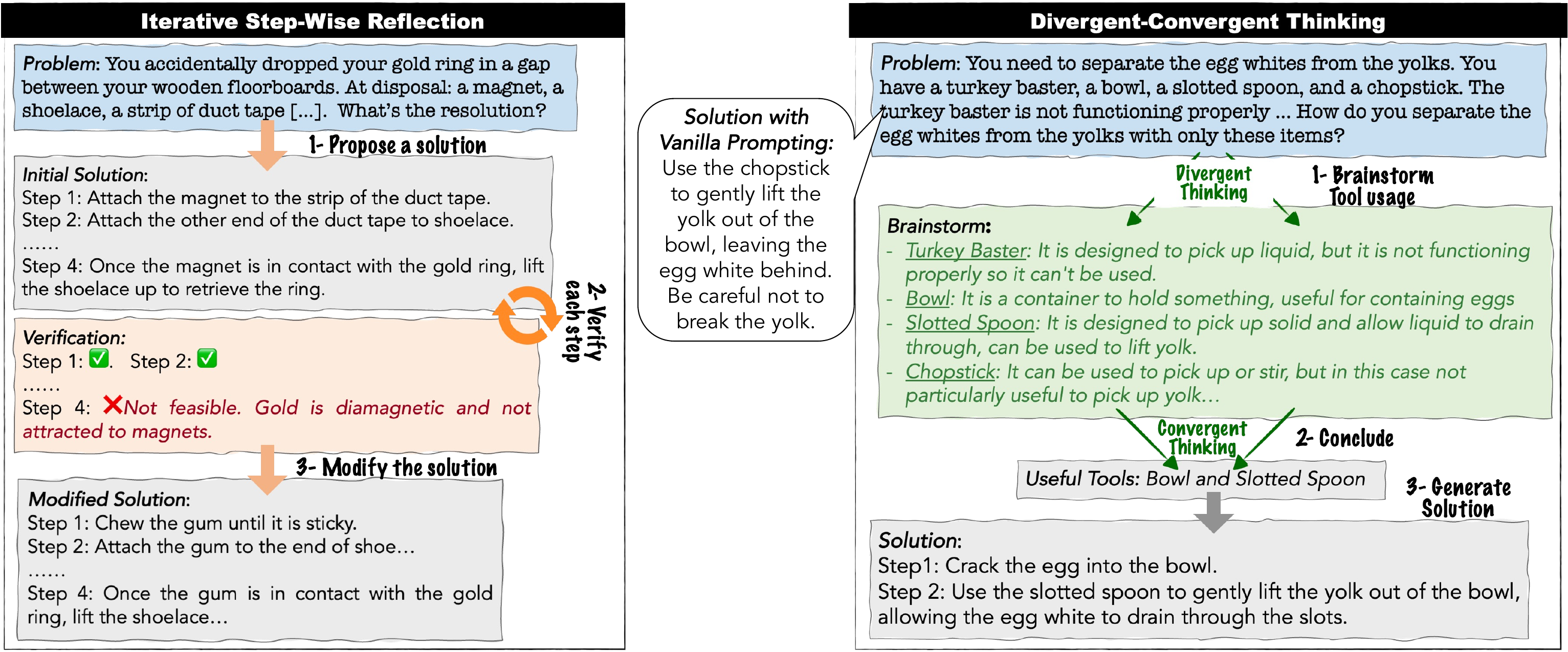}
    \vspace{-2mm}
    \caption{
    Proposed prompting methods: iterative step-wise reflection (left), divergent-convergent thinking (right).
}
    \label{fig:prompting_illustration}
    \vspace{-4mm}
\end{figure*}

We find that GPT-4 is highly prone to proposing
\textit{\textbf{physically infeasible, unwanted, or wrong actions}}. In Table \ref{table:error_analysis}, error type \textit{(1) wrong tool usage} accounts for $\sim$half of all the errors made (42.4\%), followed by \textit{(2) not achieving the goal} (17.7\%). It is crucial to highlight that \textit{\textbf{LLMs act in a fictional setting}}, failing to realize the consequences of their proposed actions and the affordances of tools in the given unconventional context. While one can argue that LLMs lack direct interaction with the physical world, the human solvers similarly contemplate the same task purely in their minds, without any visual or physical cues. 
 We also observe \textit{\textbf{two types of hallucination}}: \textit{(3) using unavailable tools} and \textit{(5) unfaithful to constraints}, which account for 16.9\% + 9.5\% = 26.4\% of all the errors made.

\subsection{Improving LLMs 
 via Prompting}\label{subsec:improve_via_prompting}

The common error types in Table \ref{table:error_analysis} motivates us to explore techniques to enhance LLMs' problem solving abilities.
Specifically, we explore two prompting strategies as illustrated in Figure \ref{fig:prompting_illustration}:
\begin{itemize}[topsep=0pt, itemsep=-2pt, leftmargin=*]
    \item \textbf{Iterative Step-Wise Reflection}
    : A self-reflection-based strategy. After the LLM generates an initial solution, we prompt it to \textit{verify} if each step is physically feasible and afforded. Subsequently, it modifies the original solution iteratively until no more modifications are needed.
    \item \textbf{Divergent-Convergent Thinking}: A cognitive-science-inspired strategy. The LLM is prompted to first enumerate the affordance of each object (\textit{i.e.,} divergent thinking) and conclude whether they are useful, followed by generating the steps towards the goal (\textit{i.e.,} convergent thinking).
\end{itemize}


\begin{figure}[t!]
\centering

    \includegraphics[width=0.45\textwidth]{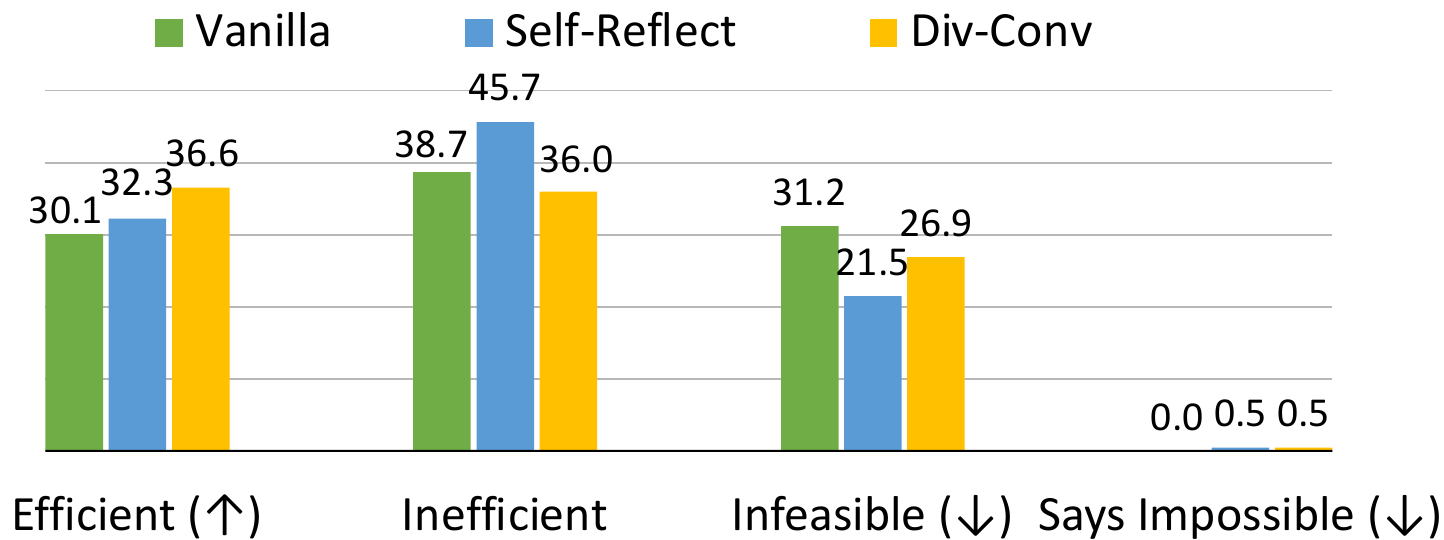}
    \vspace{-2mm}
    \caption{Results of different prompting strategies with GPT-4 
    in a zero-shot fashion: 1) vanilla prompting, 2) iterative step-wise reflection (self-reflect), and 3) divergent-convergent thinking (div-conv). 
    }
    \label{table:improve_prompting_result}
    \vspace{-3mm}
\end{figure}

We implement both prompting strategies with GPT-4, Claude2, and Llama2-13b on 180 randomly-sampled  solvable problems that do not overlap with those used in \cref{subsec:error_analysis}. The performance of the standard prompting and two proposed improvements for GPT-4 (and the remaining two LLMs) are shown in Figure \ref{table:improve_prompting_result} (and Appendix \ref{appendix:additional_prompting_results}).

For GPT-4, both proposed prompting methods contribute to a reduction in infeasible solutions. Intuitively,
\texttt{Self-Reflect}, which is designed to verify the feasibility of steps, has a larger improvement in reducing infeasible solutions (9.7\% vs 4.3\% drop); while \texttt{Div-Conv Thinking}, which is designed for better preparation before generating the solution, is more helpful in generating efficient solutions (6.5\% vs 2.2\% gain). Comparing all three LLMs, \texttt{Div-Conv Thinking} is shown to be beneficial for all, both in terms of efficiency and feasibility, but Claude2 and Llama2’s performances do not improve with \texttt{Self-Reflect}. Such a finding implies that, smaller models so far still lack the inherent ability to self-reflect and reason about physical consequences which GPT-4 is capable of.

\section{Related Work}
\paragraph{Creativity Theory}

\begin{table}[!t]
\centering\scriptsize
\begin{tabular}{@{}lll@{}}
\toprule
Model & Description            & Example                                                                                                         \\ \midrule
\texttt{Mini-C} &
  \begin{tabular}[c]{@{}l@{}}Developmental achievement\\ in the learning process.\end{tabular} &
  \begin{tabular}[c]{@{}l@{}}A pupil applying a strategy \\ learned in a math class into \\ her science project.\end{tabular} \\
  \rowcolor[HTML]{EFEFEF}
\texttt{Little-C} &
  \begin{tabular}[c]{@{}l@{}}Everyday innovation that\\ ordinary people engage with.\end{tabular} &
  \begin{tabular}[c]{@{}l@{}}Removing wrinkles on a\\ shirt without possession of \\ an iron.\end{tabular} \\
\texttt{Pro-C} & Professional expertise & \begin{tabular}[c]{@{}l@{}}Writing poems or stories\\ that receive professional\\ recognition.\end{tabular} \\
\rowcolor[HTML]{EFEFEF}
\texttt{Big-C} &
  \begin{tabular}[c]{@{}l@{}}Legendary innovation that\\ redirect a field.\end{tabular} &
  \begin{tabular}[c]{@{}l@{}}Albert Einstein arriving at\\ general relativity.\end{tabular} \\ \bottomrule
\end{tabular}
\caption{The Four-C model of creativity.}
\label{table:four-c-model}
\vspace{-3mm}
\end{table}

\citet{guilford1967creativity} defines a meaningful creative process as an interplay between spontaneous (divergent, to come up with novel ideas) and controlled (convergent, to satisfy the demand of the task) modes of thinking. 
 \citet{kaufman2009beyond} categorize human creative activities into four dimensions in Table \ref{table:four-c-model}, ranging from everyday innovation that ordinary people have knowledge of (\textit{e.g.,} removing wrinkles on a shirt without possession of an iron) to highly eminent innovation that few people engage with. 

In the AI-related creativity community, everyday innovation which better reflects the activities that most people may engage in, is under-explored possibly due to the lack of a sizable dataset. For example, \citet{koivisto2023best} study problems with four objects: rope, box, pencil, and candle. We bridge this gap by contributing a dataset with 1,600 everyday problems.

\paragraph{Cognitive Bias}
\textit{Functional fixedness} is a cognitive bias limiting our ability to use familiar objects in novel ways. For example, struggling to see a chair as anything other than a seat exemplifies this. These biases subtly impact our daily decisions, often unconsciously.
Over 82\% of the solvable problems in \DatasetName{} require using tools unconventionally to bypass such a bias. A similar work to ours \cite{collins2022structured} explored LLMs' problem-solving ability in out-of-distribution reasoning tasks.

\paragraph{Machine Physical Reasoning}

Previous research such as \citet{DBLP:journals/corr/abs-2112-05136} and \citet{NEURIPS2019_4191ef5f} investigated physical reasoning in visual contexts.
In the realm of language-based physical reasoning, prior studies primarily focused on understanding physical concepts and attributes of various objects, such as PROST \cite{aroca2021prost}, and NEWTON \cite{wang2023newton}. Relatedly, SWAG \cite{zellers-etal-2018-swag} introduced the task of grounded commonsense inference about physical situations. PIQA \cite{bisk2020piqa}, which tests machines' physical commonsense reasoning ability is most similar.
While proficiency in addressing problems in \DatasetName{} involves all the above abilities, our emphasis extends beyond.
We focus on  unconventional tool usage, reasoning over the affordance of tools and ruling out unnecessary ones, and how individual objects can be used in combination to achieve a complex goal.

\section{Discussion and Conclusion}
\paragraph{Significance of Work} We propose a new playground and the accompanying \DatasetName{} dataset for creative problem solving, which covers a broad range of topics for \textit{everyday innovation}, such as household, training, and outdoor sports, which is \textit{orthogonal} to the existing areas of reasoning and creativity, and adds to the  spectrum of machine intelligence.

The area of daily innovation, or “little-c” according to the creativity theory (Table \ref{table:four-c-model}), is a \textit{stand-alone type} of creativity and better reflects the creative activities that normal people engage with, but is much less studied than math, logical reasoning, or writing problems. These so-called daily activities can be complex too, by involving multiple-step planning for efficiency, ruling out possibilities in a large search space, using multiple tools in an unconventional manner that even humans find difficult. Namely, solving these ``daily activities" requires different kinds of creativity from scientific discovery, art, \textit{etc.}, and have a high potential for AI making people’s daily life more enjoyable.

\paragraph{Conclusion} We present \DatasetName{}, a novel benchmark focusing on everyday innovation that is carefully collected with quality and diversity control. We evaluate and compare both LLM and human performances, and highlight failure modes of LLMs in proposing physically feasible actions towards a goal. Nonetheless, we find LLM capabilities to be complementary to human capabilities under certain domain-specific settings. We propose two new prompting methods that effectively improve this reasoning ability in LLMs.

\section{Future Opportunities}
We hope \DatasetName{} dataset opens the door to multiple future directions that will contribute to the broader goal of creating \textit{AI systems that can intelligently and flexibly interact with their surroundings}. For example in this paper, we provide a preliminary attempt to improve the capability of LLMs via two prompting strategies. We encourage future investigation into planning and reasoning strategies to enhance LLMs with physical knowledge and spatial understanding, and to reduce hallucination. To further ameliorate the mistakes made by LLMs in a fictional setting, future work are encouraged to build embodied agents that can interact with physical or simulated worlds and receive feedback from the environment. 

Finally, we encourage automatic evaluation methods for this complex reasoning task. For example, using LLMs to extract claims from the candidate solutions, and examine the physical feasibility (or predict the consequences) of proposed actions based on some physical world knowledge.

\section*{Acknowledgements}

This work was funded in part by DARPA MCS program through NIWC Pacific (N66001-19-2- 4031), the NOMIS Foundation, and the Allen Institute for AI. We thank Jena D. Hwang, Ilia Sucholutsky, Mosaic team members, and the anonymous reviewers for the helpful discussions. 
\section*{Limitations} 

Measuring how well a model can solve creative problems is hard due to the lack of standardized automated metrics. For example, assuming the availability of multiple references, popular automatic NLG metrics exhibit a weak correlation with human judgment, with Pearson correlation coefficients of 0.07 for BLEU-2/BLEU-3 \cite{papineni-etal-2002-bleu} and 0.12 for BertScore \cite{zhang2019bertscore}. Our experiments thus rely on human evaluation process, which is relatively slow and costly. Therefore, new proposals for efficient and automatic evaluation framework for creative and sequential planning could be a compelling future direction. In addition to the \DatasetName{} Dataset, we release human annotations for all the solutions tested in benchmarking. We hope these additional 4,100 answer-annotation pairs, containing a full gradient of correctness (completely wrong, partially correct, correct but less efficient, and perfect), will facilitate future works in automatic evaluation.

Another limitation of our study lies in the nature of our problems being generated by an LLM, GPT-4. Despite its strengths in exploring a unique and novel angle of problem-solving, it might also exhibit inherent biases and tendencies of the underlying model. Given GPT-4's predominant training on English-speaking data, we may inadvertently reflect the cultural nuances of North American and European contexts. 
\newpage
\bibliography{anthology,custom}
\newpage
\appendix

\clearpage
\newpage\section{Additional Results}

\subsection{Comparing GPT-4 with Humans}\label{appendix:subsec:compare}
\begin{figure}[t]
    \centering
    \includegraphics[width=0.85\linewidth]{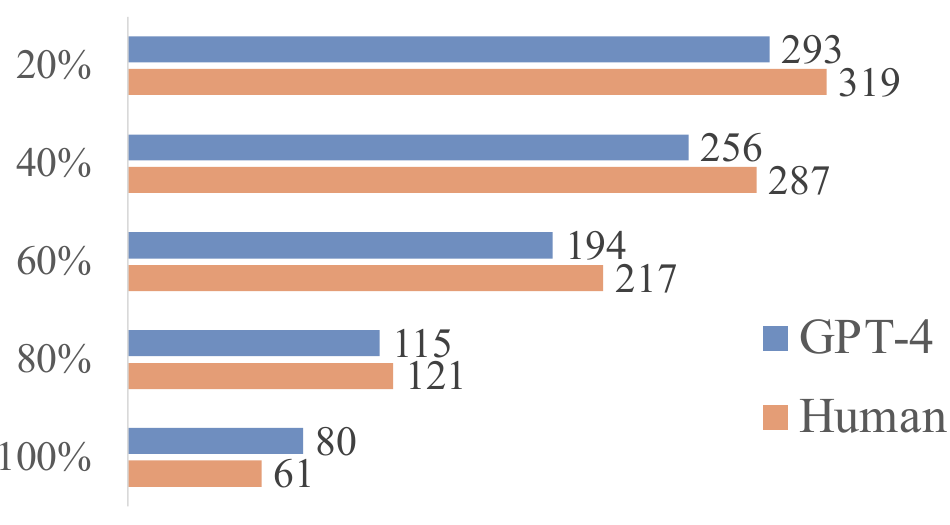}
    \caption{Number of problems (out of 323) that at least 20\%, 40\%, 60\%, 80\%, 100\% human participants (or GPT-4) answer correctly.}
    \label{fig:percent_compare}
\end{figure}

\paragraph{What percentage of individual humans outperform AI?} Figure \ref{fig:percent_compare} compares human and machine by showing the number of problems (out of 323) that at least 20\%, 40\%, 60\%, 80\%, and 100\% human participants (or GPT-4) answer correctly. Given the unique strengths and knowledge scopes of different individuals, it is less likely that all human participants can answer the same problem correctly. However, there is a higher chance where at least 60\% human participants know the answer.

\begin{figure}[h]
\centering
    \includegraphics[width=0.45\textwidth]{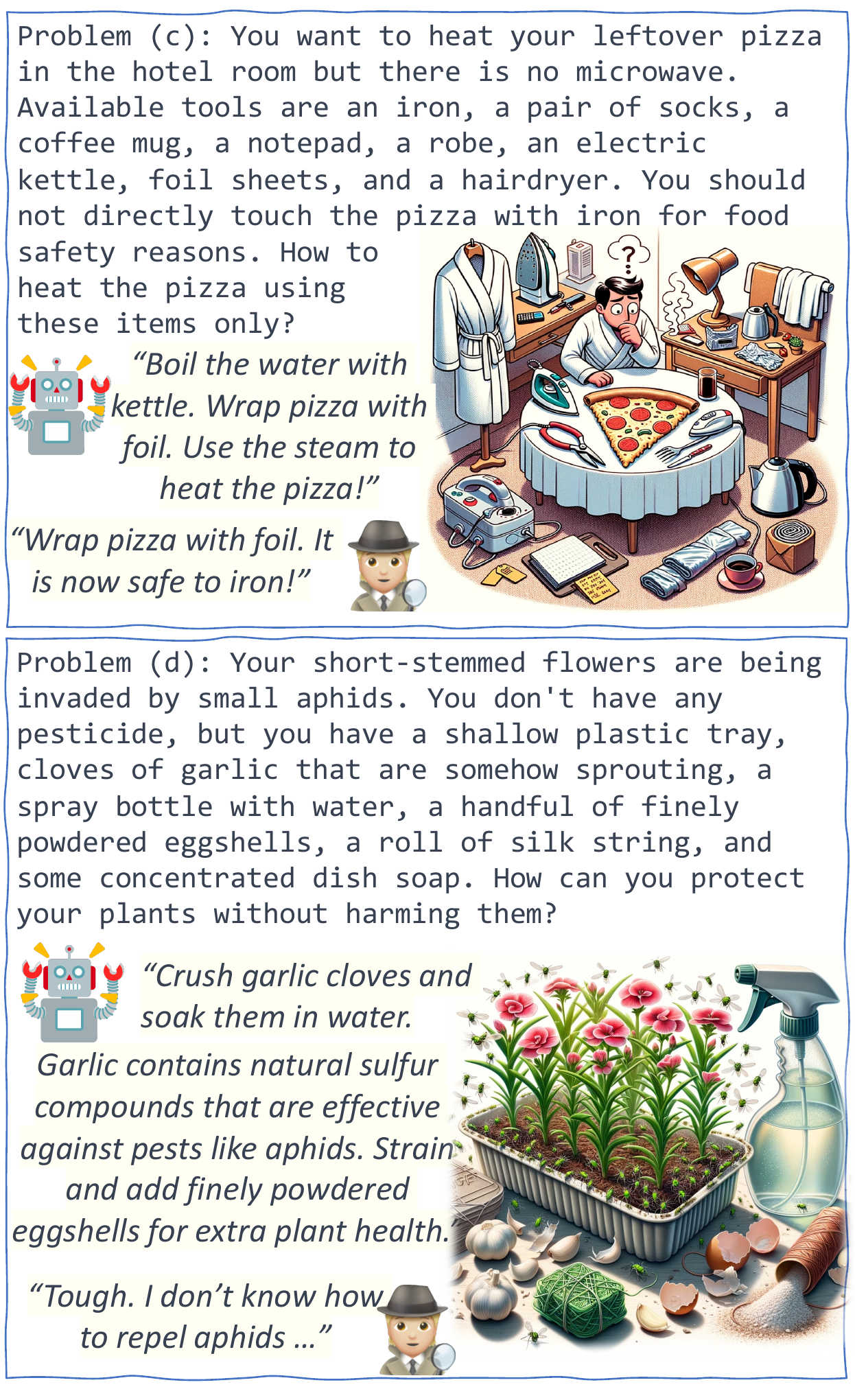}
    \vspace{-3mm}
    \caption{Detailed examples showing the complementary capabilities of human and GPT-4. In problem (c), human participants find a more efficient solution to heat the pizza than AI. In problem (d), humans fail to solve this highly-specialized task to repel aphids, whereas the LLM has equipped itself with domain knowledge on gardening during massive pre-training.
    }
    \vspace{-3mm}
    \label{fig:teaser_more}
\end{figure}

\paragraph{Complementary capabilities of human and AI.} Figure \ref{fig:teaser_more} presents two examples showing the complementary capabilities of human and AI in creative problem solving. In problem (c), human participants find a more efficient solution to heat the pizza than GPT-4. In problem (d) requiring domain knowledge gardening, humans fail to solve this highly-specialized task, whereas the LLM has equipped itself with such domain knowledge during massive pre-training.

\paragraph{What tools are human more proficient at?}
Recall that in \cref{subsec:compare_2d_plot} we convert the categorical labels into numerical scores ranging from 0 (Fail) to 1 (Perfect) to conduct problem-wise analysis. Similarly, we conduct object-wise analysis by first parsing the tools presented in each problem, and then calculating the same numerical scores for each tool. Note that we opt to parse all the tools presented in the problem setting instead of those actually used in a proposed solution, because being able to reason about the potential usage of presented tools and conclude to \textit{not} use a possible tool is also an keystone towards intelligence. We identify several tools that humans and GPT-4 attempt most differently and report them in Table \ref{table:compare_difference_tools}. For example, humans are more proficient at attempting magnifying glass, rocks, calculators, knifes, etc., whereas AIs are better attempting mirrors, gloves, and scarves. In general, there are more tools humans are proficient at. 

\begin{table}[h]
\small\centering
\begin{tabular}{@{}cll@{}}
\toprule
                                           & Object              & \begin{tabular}[c]{@{}l@{}}Human-AI\\ Difference\end{tabular}   \\\midrule
                                
\multirow{11}{*}{A. human\textgreater{}AI} & magnifying glass & 0.602                                                         \\
                                           & rock             & 0.447                                                         \\
                                           & calculator       & 0.405                                                         \\
                                           & kitchen knife    & 0.386                                                         \\
                                           & hair tie         & 0.359                                                         \\
                                           & paper cup        & 0.292                                                         \\
                                           & zip ties         & 0.283                                                         \\
                                           & pen              & 0.281                                                         \\
                                           & kettle           & 0.273                                                         \\
                                           & old t-shirt      & 0.252                                                         \\
                                           & sunscreen        & 0.25                                                          \\ \midrule
\multirow{5}{*}{B. human\textless{}AI}     & mirror           & -0.314                                                        \\
                                           & gardening gloves & -0.311                                                        \\
                                           & scarf            & -0.307                                                        \\
                                           & tablecloth       & -0.289                                                        \\
                                           & clothespins      & -0.253                                                        \\ \bottomrule
\end{tabular}
\caption{Tools that human are more proficient at leveraging or deciding to not leverage than AI (GPT-4 in our case), and vice versa.}
\label{table:compare_difference_tools}
\end{table}

\subsection{Benchmark Results}\label{appendix:benchmark}
\newcolumntype{?}{!{\vrule width 1pt}}
\begin{table*}[t!]
\footnotesize\centering
\renewcommand{\arraystretch}{1.0}
\begin{tabular}{l?ccc|c?ccc|c} \toprule
 &
  \multicolumn{3}{c}{\cellcolor[HTML]{C6E0B4}\textbf{Correct (\%)}} &
  \cellcolor[HTML]{C6E0B4} &
  \multicolumn{3}{c}{\cellcolor[HTML]{F4B084}\textbf{Wrong (\%)}} &
  \multicolumn{1}{c}{\cellcolor[HTML]{F4B084}} \\
\textbf{} &
  \cellcolor[HTML]{E2EFDA}\textbf{\begin{tabular}[c]{@{}c@{}}A. Eff-\\icient\end{tabular}} &
  \cellcolor[HTML]{E2EFDA}\textbf{\begin{tabular}[c]{@{}c@{}}B. Less\\ Efficient\end{tabular}} &
  \cellcolor[HTML]{E2EFDA}\textbf{\begin{tabular}[c]{@{}c@{}}C. Uns-\\ olvable\end{tabular}} &
  \multirow{-2}{*}{\cellcolor[HTML]{C6E0B4}\textbf{\begin{tabular}[c]{@{}c@{}}Correct in \\ Total  (↑)\end{tabular}}} &
  \cellcolor[HTML]{FCE4D6}\textbf{D. Partial} &
  \cellcolor[HTML]{FCE4D6}\textbf{E. Mostly} &
  \cellcolor[HTML]{FCE4D6}\textbf{\begin{tabular}[c]{@{}c@{}}F. Fail to\\ Identify\end{tabular}} &
  \multicolumn{1}{c}{\multirow{-2}{*}{\cellcolor[HTML]{F4B084}\textbf{\begin{tabular}[c]{@{}c@{}}Wrong in\\ Total (↓)\end{tabular}}}} \\ \hline

  \multicolumn{9}{c}{\texttt{\scriptsize{Single Effort}}}  \\ \hline
\textbf{Llama2-7b} & 8.9  & 18.1 & 8.5  & 35.5 & 6.9  & 27   & 30.6 & 64.5 \\
\textbf{Llama2-13b} & 11.7 & 28   & 2.3  & 42.0   & 12.1 & 32.3 & 13.6 & 58.0   \\
\textbf{Llama2-70b} & 11.6 & 24   & 5.6  & 41.2 & 14.0   & 27.2 & 17.6 & 58.8 \\
\textbf{PaLM2} & 14.7 & 25.9 & 0.0    & 40.6 & 10.8 & 35.5 & 13.1 & 59.4 \\
\textbf{Claude2} & 14.0   & 22.2 & 16.5 & 52.7 & 8.2  & 12.3 & 26.7 & 47.2 \\
\textbf{GPT-3.5} & 13.8 & 15.4 & 11.4 & 40.6 & 10.2 & 11.4 & 37.8 & 59.4 \\
\textbf{GPT-4 (Random)} & 24.8 & 35.5 & 2.1  & {\ul {62.4}}    & 11.9 & 14.9 & 10.8 & {\ul {37.6}}   \\
\textbf{Human (Random)} & 27.6 & 27.6 & 9.9  & \textbf{65.1} & 5.6  & 10.8 & 18.6 & \textbf{35.0}  \\ \hline
  \multicolumn{9}{c}{\texttt{\scriptsize{Multiple Efforts}}}  \\ \hline
{\color[HTML]{757171} \textbf{Average   GPT-4}} &
  {\color[HTML]{757171} 24.8} &
  {\color[HTML]{757171} 33.2} &
  {\color[HTML]{757171} 5.0} &
  {\color[HTML]{757171} 63.0} &
  {\color[HTML]{757171} 12.5} &
  {\color[HTML]{757171} 15.7} &
  {\color[HTML]{757171} 8.7} &
  {\color[HTML]{757171} 36.9} \\
{\color[HTML]{757171} \textbf{Average Human}} &
  {\color[HTML]{757171} 26.2} &
  {\color[HTML]{757171} 28.7} &
  {\color[HTML]{757171} 12.9} &
  {\color[HTML]{757171} 67.8} &
  {\color[HTML]{757171} 5.1} &
  {\color[HTML]{757171} 10.2} &
  {\color[HTML]{757171} 16.9} &
  {\color[HTML]{757171} 32.2} \\

\cellcolor[HTML]{E7E6E6}\textbf{Best GPT-4} &
  \cellcolor[HTML]{E2EFDA}62.5 &
  \cellcolor[HTML]{E2EFDA}21.1 &
  \cellcolor[HTML]{E2EFDA}8.7 &
  \cellcolor[HTML]{E2EFDA}{\ul {92.3}} &
  \cellcolor[HTML]{FCE4D6}2.2 &
  \cellcolor[HTML]{FCE4D6}4.3 &
  \cellcolor[HTML]{FCE4D6}1.2 &
  \cellcolor[HTML]{FCE4D6}{\ul {7.7} }\\
\cellcolor[HTML]{E7E6E6}\textbf{Best Human} &
  \cellcolor[HTML]{A9D08E}72.8 &
  \cellcolor[HTML]{A9D08E}15.2 &
  \cellcolor[HTML]{A9D08E}10.8 &
  \cellcolor[HTML]{A9D08E}\textbf{98.8} &
  \cellcolor[HTML]{F4B084}0.6 &
  \cellcolor[HTML]{F4B084}0.6 &
  \cellcolor[HTML]{F4B084}0.0 &
  \cellcolor[HTML]{F4B084}\textbf{1.2}\\ \bottomrule
\end{tabular}
\vspace{-2mm}
\caption{\textbf{Top}: Benchmark results of seven LLMs and human with a single effort. 
For human participants, there is no single participant who worked on all problems. So we take a random response from each problem. \textbf{Bottom}: Comparison between GPT-4 and human where we evaluated multiple solutions per problem. The best performance, which can be viewed as an upper bound, is computed by taking the individual best answer (out of 4) for each problem.  
We use boldface to denote the best performance and underline to denote the second best. 
}
\vspace{-3mm}
\label{Table:benchmark-results}
\end{table*}

We report the benchmark results in Table \ref{Table:benchmark-results}. Category \textbf{A}, \textbf{B}, and \textbf{C} are the three aspects of correct responses, while the remaining \textbf{D}, \textbf{E}, and \textbf{F} are aspects of the wrong ones. At a glance, despite varying in their characteristics, all of the benchmarked LLMs lag behind the performance of humans.

\subsection{Enhancing LLMs' Problem Solving}

\paragraph{Results with Claude2 and Llama2}\label{appendix:additional_prompting_results}

\begin{figure}[t]
\centering
    \includegraphics[width=0.5\textwidth]{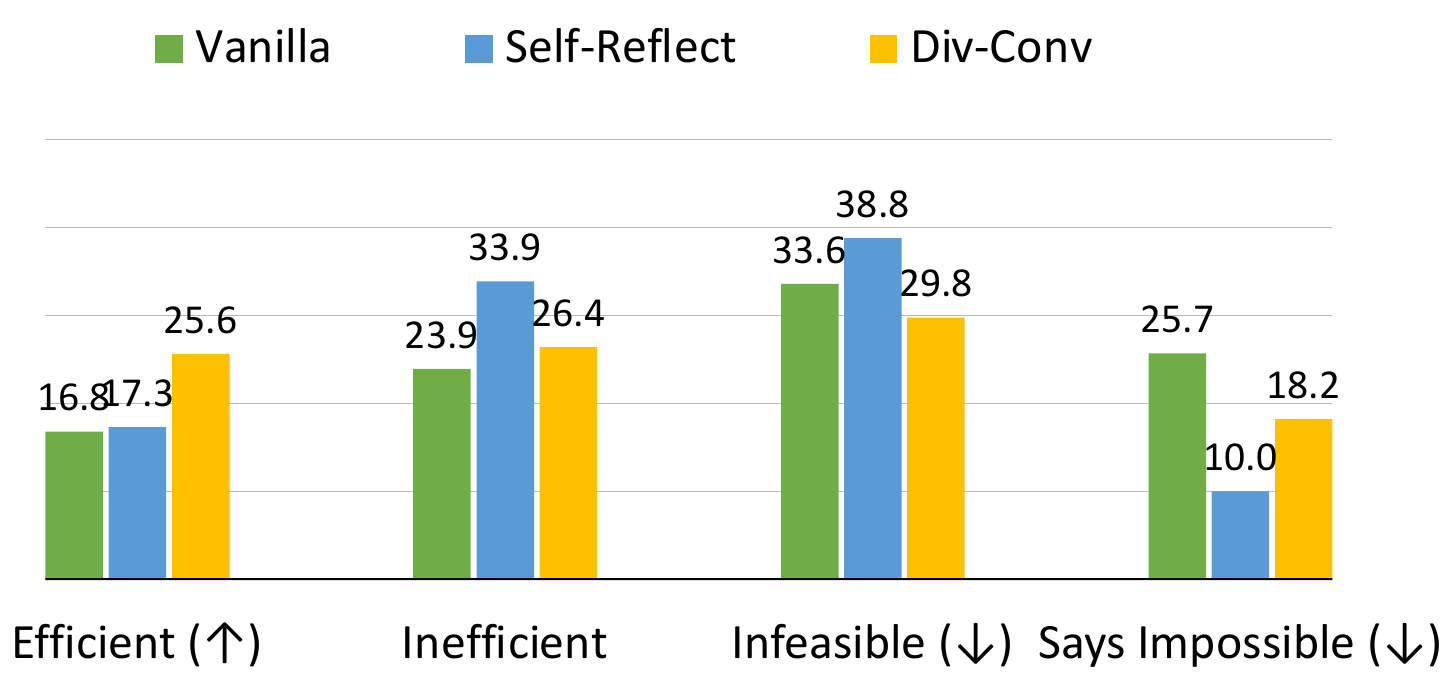}
    \caption{Results of different prompting strategies with Claude2. We compare 1) vanilla prompting, 2) iterative step-wise reflection (reflect), and 3) divergent-convergent thinking (div-conv).
    }
    \label{table:improve_prompting_result_claude2}
\end{figure}

\begin{figure}[t]
\centering
    \includegraphics[width=0.5\textwidth]{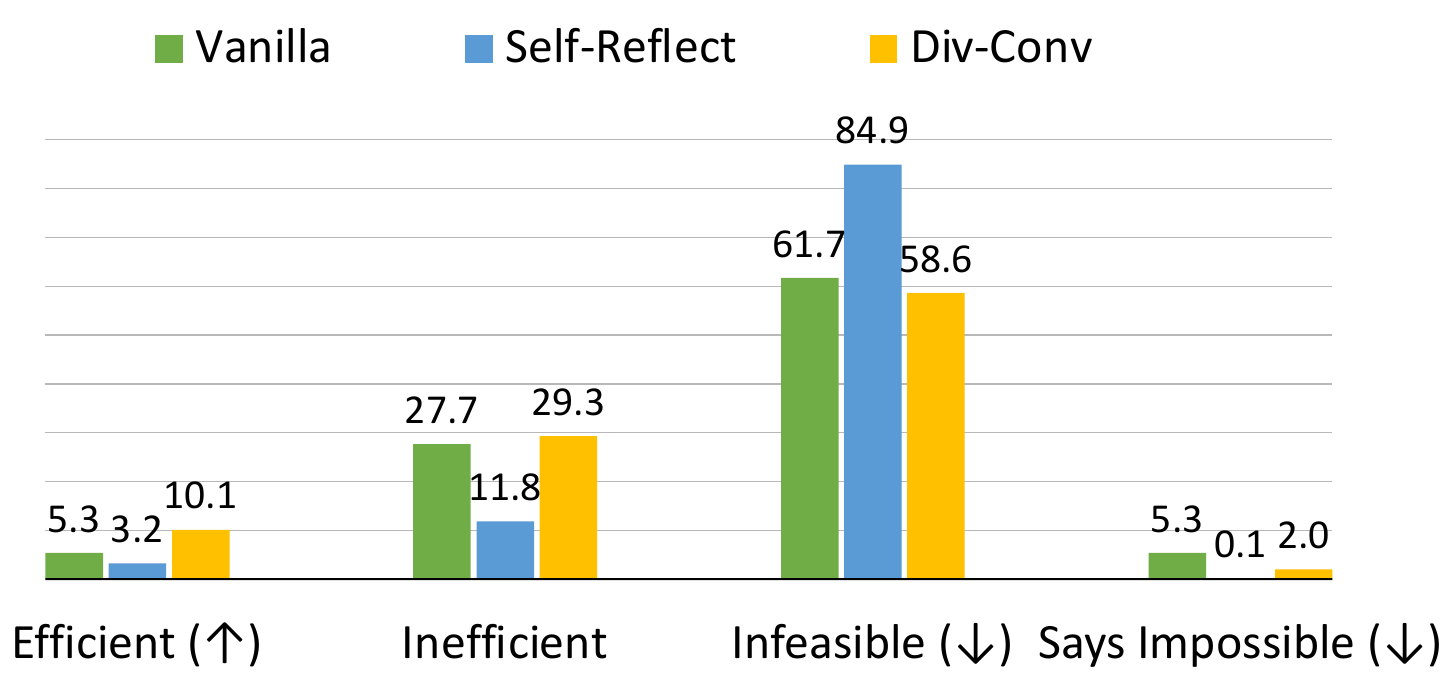}
    \caption{Results of different prompting strategies with Llama2-13b. We compare 1) vanilla prompting, 2) iterative step-wise reflection (reflect), and 3) divergent-convergent thinking (div-conv).
    }
    \label{table:improve_prompting_result_llama2}
\end{figure}

We report the performance of the standard, zero-shot prompting and two proposed improvements for Claude2 and Llama2-13b in Figure \ref{table:improve_prompting_result_claude2} and Figure \ref{table:improve_prompting_result_llama2}. Different from GPT-4 (shown in Figure \ref{table:improve_prompting_result}), the \texttt{self-reflection} strategy does not help any of these two models to reduce infeasible answers. When prompted to reflect on its previous answer, Llama2 always claims that its original answer is mistaken and attempts to correct itself blindly. We hypothesize that these two LLMs are weaker than GPT-4 and lack the inherent ability to faithfully conduct complicated physical reasoning. On the other hand, we see that \texttt{Divergent-Convergent Thinking} is beneficial for all LLMs across all dimensions.

\section{The Four-C Creativity Model}\label{appendix:four_c_model}

 \citet{kaufman2009beyond} propose the Four-C model (Table \ref{table:four-c-model}), categorizing human creative activities into \texttt{Mini-C}: developmental creativity in the learning process, \texttt{Little-C}: everyday innovation that ordinary people have knowledge of and engage with (such as removing wrinkles on a shirt without possession of an iron)
 , \texttt{Pro-C}: professional expertise such as writing poems or painting artwork, and \texttt{Big-C}: highly eminent innovation that few people engage with.

\begin{table*}[t]
\centering
\small
\begin{tabular}{@{}lll@{}}
\toprule
\textbf{Indoors/Household} & \textbf{Neutral}                 & \textbf{Outdoors}              \\ \midrule
bedroom                    & at a party                       & at the beach                   \\
closet or storage organization & classroom and university lecture hall & backyard gardening                        \\
cooking a complex dish         & dog training                          & beach cleanups, or planning a beach event \\
dining room                & garage                           & boat trip                      \\
fitness workouts           & going out for a meal             & campsite setting               \\
gym and sports facilities      & plants, flowers and garden            & city streets and sidewalks                \\
hair styling and care      & public speaking                  & construction work              \\
home improvement           & recycling and waste management   & desert survival                \\
in a hotel room            & school and student activity      & exploring a cave               \\
indoors arrangement        & school science fair              & farm duties                    \\
kitchen                    & science laboratory               & forest and jungle              \\
library                    & swimming                         & hiking, camping, and traveling \\
living room                & university campus                & in the parks                   \\
office and work            & vehicle maintenance              & in the rain                    \\
packing things up          & weather preparation and response & in the winter                  \\
personal   grooming and beauty routine   &     & in the zoo                                \\
shopping                   &                                  & on the playground              \\
                           &                                  & playing with snow              \\
                           &                                  & playing with water             \\
                           &                                  & rooftop terrace                \\ \bottomrule
\end{tabular}
\caption{The tags (\textit{i.e.,} locations and activities) used to curate the dataset for diversity control. They can be broadly divided into Indoors/Household, Neutral, and Outdoors.}
\label{table:tags}
\end{table*}
\section{More Information on the \DatasetName{} Dataset}\label{appendix:dataset_details}

\subsection{Human Verification Process}

After generating the challenging scenarios, we involve human verifiers to judge if the final versions of the problems \textbf{1)} are solvable (\textit{i.e.,} it is possible to find a reasonable solution using the presented tools), unsolvable, or need more clarification (\textit{i.e.,} the setup is vague or contradictory, which will be discarded), and \textbf{2)} for those solvable, whether solving them efficiently requires creative thinking (\textit{i.e.,} using objects to achieve goals they were \textit{not} originally designed for \textemdash unconventional usage). 
Each problem is annotated by three human verifiers from Amazon Mechanical Turk.  The detailed verification interface can be found in Appendix \ref{appendix:survey}. The average inter-annotator agreement (IAA), measured by Cohen's Kappa,  are 0.67 and 0.77 for tasks \textbf{1)} and \textbf{2)}, respectively. 

\subsection{Collecting Gold Solutions}\label{subsec:collect_gold}

We provide more details on the final step of our data collection \textemdash to pair each problem with a gold answer. For the solvable subset, the answer is a feasible solution written step by step. For the unsolvable subset, the answer is a correct explanation for why the stated goal cannot be achieved. 

To save human effort, we start by leveraging the generative strengths of a powerful LLM, \textit{i.e., }GPT-4. Specifically, we first prompt GPT-4 to generate a solution for each problem in the \DatasetName{} dataset. Then, human verifiers assess whether the generated solutions are valid. Only if \textbf{all three verifiers} agree that a solution is valid, it becomes part of our dataset. Otherwise, we ask human workers to write down a solution (for solvable subset) or a justification (for unsolvable subset).

\subsection{Does the data collection pipeline result in progressively challenging problems?}\label{subsec:result_3_iterations}

To test whether our data creation pipeline (in Figure \ref{fig:progressive_data_creation}) is indeed iteratively posing challenge to a previous iteration, we collect GPT-4 answers to iteration 1, 2, and 3 of 200 problems, and run the same human evaluation process described in \cref{subsec:human_eval}.

GPT-4's performance on all three iterations of the same set of problems can be found in Table \ref{table:3_iterations}. As the problems get iteratively refined, the ratio of feasible and efficient solutions decrease, and the ratio of infeasible answers increase. This reflects that most potent LLM, GPT-4, indeed finds the problems increasingly challenging.

\begin{table}[h]
\small\centering
\renewcommand{\arraystretch}{1.2}
\setlength\tabcolsep{3pt}
\begin{tabular}{@{}ccccc@{}}
\toprule
\textbf{Solutions} &
  \textbf{\begin{tabular}[l]{@{}l@{}}Feasible \\ \& eff. ($\uparrow$)\end{tabular}} &
  \textbf{\begin{tabular}[l]{@{}l@{}}Feasible \\ \& ineff.\end{tabular}} &
  \textbf{\begin{tabular}[l]{@{}c@{}}Infeasible\\ ($\downarrow$)\end{tabular}} &
  \textbf{\begin{tabular}[l]{@{}l@{}}LLM says \\ unsolv. ($\downarrow$)\end{tabular}} \\ \midrule
\rowcolor[HTML]{E7E6E6} 
Iteration 1 & 39.1\% & 36.8\% & 24.0\% & 0.1\% \\
Iteration 2 & 34.7\% & 32.2\% & 31.7\% & 1.4\% \\
\rowcolor[HTML]{E7E6E6} 
Iteration 3 & 25.4\% & 37.9\% & 35.7\% & 1.0\% \\ \bottomrule
\end{tabular}
\caption{GPT-4 performance on iteration 1, 2, and 3 of 200 problems. Numbers in each row add up too 100\%.}
\label{table:3_iterations}
\end{table}

\subsection{Diversity Control}\label{appendix:diversity_control}
\paragraph{Tags used for Diversity Control}
Before the first iteration, we hand craft more than 50 tags of locations and activities, aiming to ensure that our data collection pipeline delves into a variety of topics. The tags cover diverse range of human activities, from indoor ones such as \textit{home arrangement} and \textit{working in the office}, to outdoor ones such as \textit{hiking}, \textit{gardening}, and \textit{playing with water}. These predefined tags are integrated into the prompt that we used to query GPT-4 for problem curation at Iteration 1. We list all the tags (\textit{i.e.,} locations and activities) used to curate the dataset in Table \ref{table:tags}. They are introduced to prompt the LLM for diversity control, and can be broadly divided into Indoors/Household, Neutral, and Outdoors.

\paragraph{Generation in Batch} All problems are generated and refined in batches of 15 rather than one by one, as we find out the former results in significantly higher diversity. We then leverage a widely-used sentence transformer \cite{reimers-2020-multilingual-sentence-bert} to filter out any newly generated problem that is semantically similar to the existing ones in our database.

\paragraph{Analyzing Tool Affordance}
We leverage GPT-4 to analyze the affordance of presented tools in the \DatasetName{} dataset. Specifically, we start with a small set of hand-crafted affordance as seed. Despite being required to choose only from this fixed list of affordances, GPT-4 does not strictly follow our instruction, and sometimes returns new types that are not included in the seed list. We then gradually expand the list of affordances with newly generated ones. 

For eliciting tool affordances, we use the prompt shown in Figure \ref{code:llm_prompt_analyze_affordance}.

\begin{figure}[h]
\begin{lstlisting}
<-- Instruction. -->
You need to write the most common affordances of an item. Please choose one or more options from the following:
<-- Seed list to expand with. -->
Container/holding items, covering, heating, measuring, drawing/writing, cleaning, sitting/stepping, tying or connecting, illumination, stretching, starting fire, sealing, cutting, separation, reaching high areas, powering devices, digging, making noise, flatten, cutting, gripping things, reflecting, eaten as food.

<-- Examples. -->
Here are some examples:
 rice: eaten as food
 case: container/holding items, protection, covering
 ruler: measuring, straightening
 box: container/holding items
 pencil: drawing/writing,

<-- Actual Task. -->
Please write the common types of affordances of the following tools.

1. {Tool 1}.
...
N. {Tool N}.


\end{lstlisting}
\caption{The prompt used to analyze tool affordance. We start with a list of affordances as seed. We gradually expand our list thanks to the fact that GPT-4 does cannot strictly follow our instruction and occasionally generates other affordances not belonging the predefined set.}
\label{code:llm_prompt_analyze_affordance}
\end{figure}
\begin{table}[t]
\small\centering
\begin{tabular}{@{}lll@{}}
\toprule
\textbf{Tool}    & \textbf{Affordance}                                                                 & \textbf{Freq.} \\ \midrule
duct tape        & sealing; tying or connecting                                                        & 2.0\%              \\
\rowcolor[HTML]{E7E6E6}
plastic bag & \begin{tabular}[c]{@{}l@{}}container or holding items;\\ covering\end{tabular}       & 0.7\% \\
flashlight       & illumination                                                                        & 0.7\%              \\
\rowcolor[HTML]{E7E6E6}
aluminum foil    & covering; heating; sealing                                                          & 0.6\%              \\
hairdryer        & heating; drying; making noise                                                       & 0.5\%              \\
\rowcolor[HTML]{E7E6E6}
ruler            & measuring; straightening                                                            & 0.4\%              \\
broom            & \begin{tabular}[c]{@{}l@{}}cleaning; sweeping; reaching \\ high areas\end{tabular}  & 0.4\%              \\
\rowcolor[HTML]{E7E6E6}
spoon            & eating; stirring; measuring                                                         & 0.4\%              \\
toothbrush       & cleaning; spraying                                                                  & 0.4\%              \\
\rowcolor[HTML]{E7E6E6}
mag. glass & magnifying; starting fire                                                           & 0.4\%              \\
rope             & \begin{tabular}[c]{@{}l@{}}tying or connecting; reaching \\ high areas\end{tabular} & 0.4\%              \\
\rowcolor[HTML]{E7E6E6}
hammer      & \begin{tabular}[c]{@{}l@{}}flattening; gripping things; \\ making noise\end{tabular} & 0.3\% \\
yoga mat    & \begin{tabular}[c]{@{}l@{}}stretching; sitting/stepping;\\ covering\end{tabular}     & 0.3\% \\
\rowcolor[HTML]{E7E6E6}
towel            & wetting; covering; absorbing                                                           & 0.3\%              \\
frisbee          & playing; throwing                                                                   & 0.3\%              \\
\rowcolor[HTML]{E7E6E6}
toothpick        & cleaning; separating                                                                & 0.3\%              \\ \bottomrule
\end{tabular}
\caption{Examples of most commonly presented tools, their featured affordances, and frequency of these tools in the entire dataset. We randomly pick 16 tools from the top 40 frequent ones in the \DatasetName{} dataset. In total, more than 3,800 different tools appear in our dataset. 
}
\label{table:common_tools}
\end{table}
\paragraph{Commonly-presented tools and their frequencies}

In total, more than 3,800 different tools appear in our \DatasetName{} dataset. We list in Table \ref{table:common_tools} 16 commonly-presented tools, their featured affordances, and frequency. The number of unique tools and the long tails in distribution signify a desirable level of diversity.

\section{Experimental Details}
\subsection{Benchmark Setup}\label{appendix:experiment_setup}

\paragraph{Recruiting MTurk Evaluators} We used qualification tasks to recruit 160 qualified annotators on Mechanical Turk. They are paid over 18 USD per hour for all the evaluation and verification tasks.

\paragraph{Collecting Human Solutions on Prolific} All participants of human study provide informed consent in accordance with an approved Princeton University
institutional review board (IRB) protocol (10859). For a given problem, participants indicated whether they believed the problem is solvable, unsolvable, or required further clarification. If solvable, they provided a step-by-step solution, and otherwise they explained why the problem was unsolvable. A screenshot of the elicitation interface is shown in Figure~\ref{fig:solver_interface}. 

\paragraph{Collecting Multiple GPT-4 Responses in Benchmark}
Recall that in \cref{subsec:collect_machine_sol}, we elicit multiple solutions exclusively from the most potent LLM, GPT-4, to emulate the same setup of human study. To align with the varying number of human responses for different problems, we adjusted the quantity of collected GPT-4 answers to match that of human answers.
On average, we elicited four GPT-4 solutions per problem through separate API call. To this end, four manually-designed instructions are used to prompt GPT-4 to reduce repetition among separate sessions. For each API call, we still adopt Nucleus sampling and return the top one sequence. 

\subsection{Analyzing Results}
Each machine-generated or human-written answer is annotated by three Mturk workers, with an average IAA of 0.71 as measured by Cohen's Kappa, indicating a substantially strong agreement. Interestingly, we notice that human workers disagree more often when deciding whether a solution is efficient or inefficient. Upon further investigation, we realize this is partially due to the limitation of individual annotator's capability -- a person who is unaware of the most efficient solution might label a sub-optimal one as highly efficient. Therefore, for those generated solutions linked to solvable problems, instead of taking the \textit{majority} vote, we take the \textit{worse} labels as the golden label (\textit{e.g.,} taking \textit{`ineff.'} from  \textit{[`eff.', `ineff.', `eff.']}). For all other cases, we still take the majority votes as gold labels. We find such modification leads to a more accurate set of labels. 


\subsection{The Prompts for Improving LLM's Ability}\label{appendix:prompt}

Figure \ref{code:llm_prompt_step_verify} and Figure \ref{code:llm_prompt_div_conv} list the actual prompts for \texttt{Self-Reflection} and \texttt{Divergent-Convergent Thinking}.

\begin{figure*}[t]
\begin{lstlisting}
<-- Round 1: -->
User: {Problem Statement} 
If the problem is solvable, provide a concise solution. Use step1, step2, etc, and mention the tools to achieve each step. Use as few steps as possible and the answer should ideally be less than 100 words.
 
If you cannot find a feasible solution, just say that it is not possible and give a very short justification.

Assistant: {Answer}

<-- Round 2: -->
User: Now, please verify if each step is physically feasible and afforded. After that, modify the solution if needed.
Use the following format:
Step 1: ...
Step 2: ...
...
Conclusion 1: Whether the problem is indeed solvable given all the constraints
Conclusion 2: (If still solvable) No modification needed/Modification needed.


Modified solution: 
Assistant:  {Response and Updated solution}
<-- Repeat until no modification is needed.-->

\end{lstlisting}
\caption{Prompt used for the \texttt{step-by-step verify} strategy.}
\label{code:llm_prompt_step_verify}
\end{figure*}

\begin{figure*}[t]
\begin{lstlisting}
User: {Problem Statement} 
Give a feasible solution very concisely. Note that some tools are not useful, so please analyze the affordance of each presented object, and rule out unnecessary ones first. 


Use the following format:
1. List the affordance of presented items and whether they are useful
2. Summary: list useful tools
3. If the problem is solvable under all these constraints, write the solution. Use step1, step2, etc, and mention the tools to achieve each step. Use as few steps as possible and the answer should ideally be less than 100 words.
 
If you cannot find a feasible solution, just say that it is not possible and give a very short justification.

Assistant: {Analysis of the affordance and the main answer}


\end{lstlisting}
\caption{Prompt used for the \texttt{divergent-convergent thinking} strategy.}
\label{code:llm_prompt_div_conv}
\end{figure*}

\subsection{Human Task Interfaces}\label{appendix:survey}

\begin{figure*}[t!]
    \centering
    \includegraphics[width=0.95\linewidth]{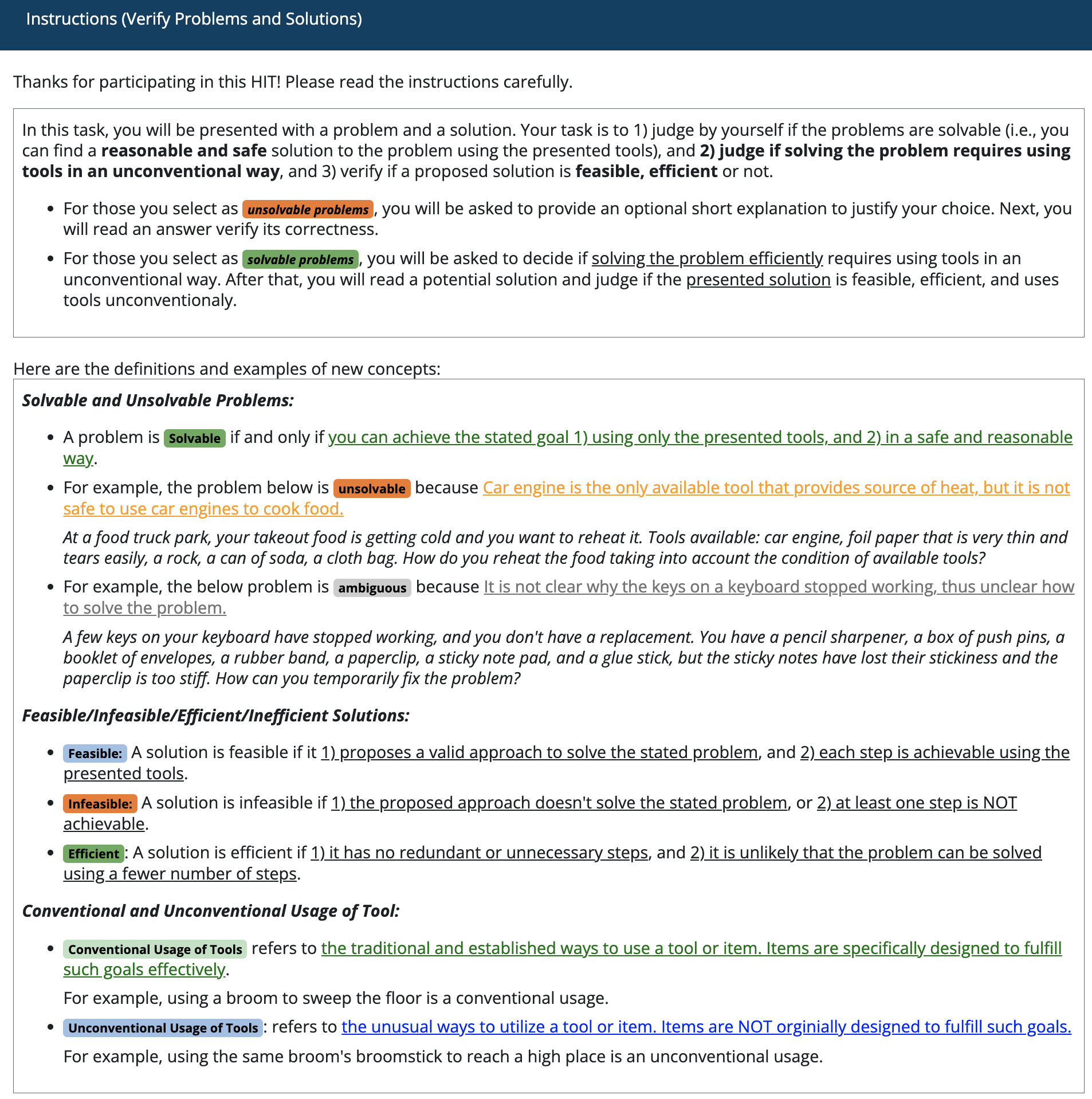}
    \vspace{-2mm}
    \caption{Human Annotation Interface for Data Collection and Difficulty Assessment, Page 1.
}
    \label{fig:survey01}
    \vspace{-5mm}
\end{figure*}

\begin{figure*}[t!]
    \centering
    \includegraphics[width=0.95\linewidth]{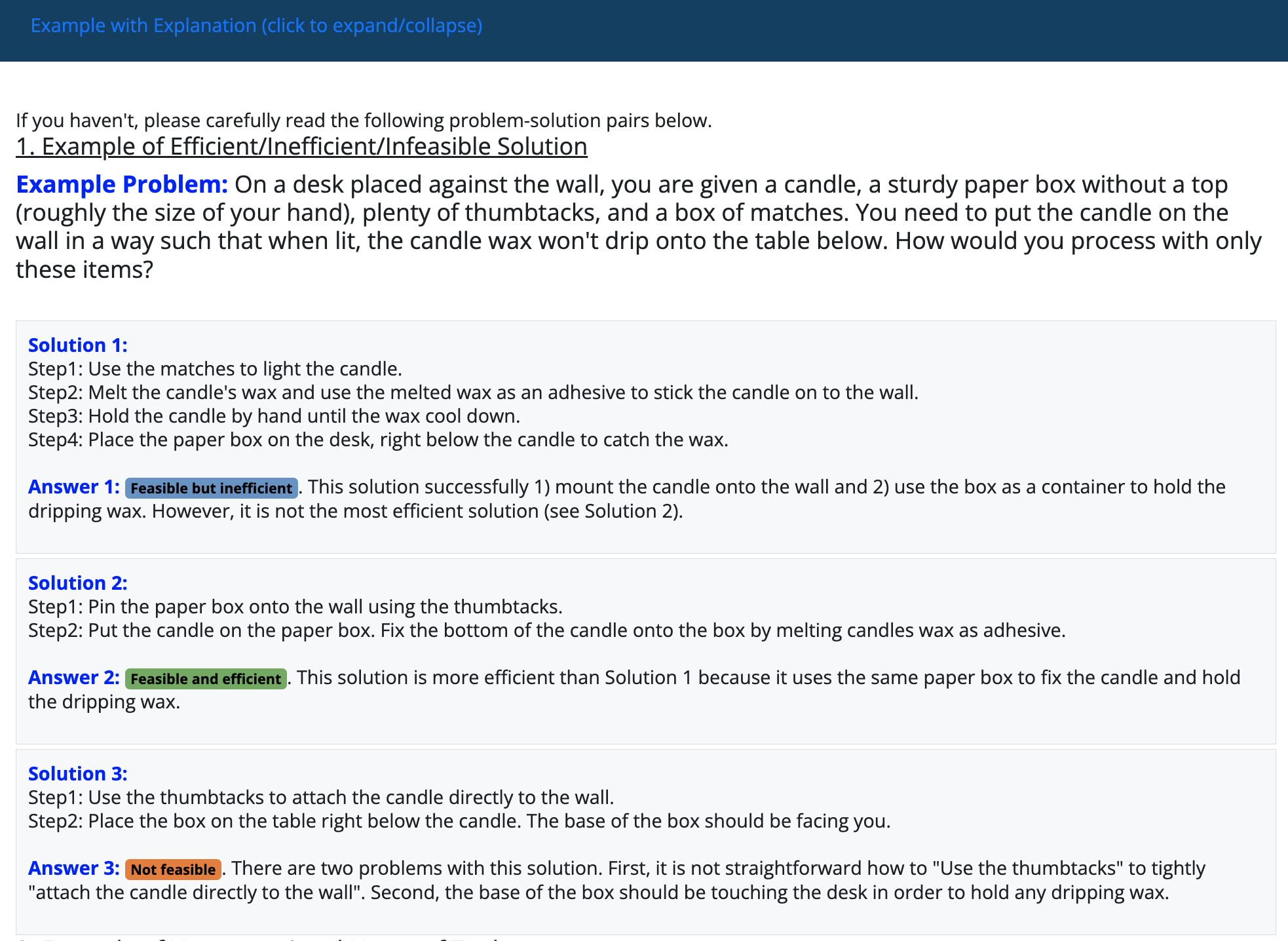}
    \vspace{-2mm}
    \caption{Human Annotation Interface for Data Collection and Difficulty Assessment, Page 2.
}
    \label{fig:survey02}
    \vspace{-5mm}
\end{figure*}

\begin{figure*}[t!]
    \centering
    \includegraphics[width=0.95\linewidth]{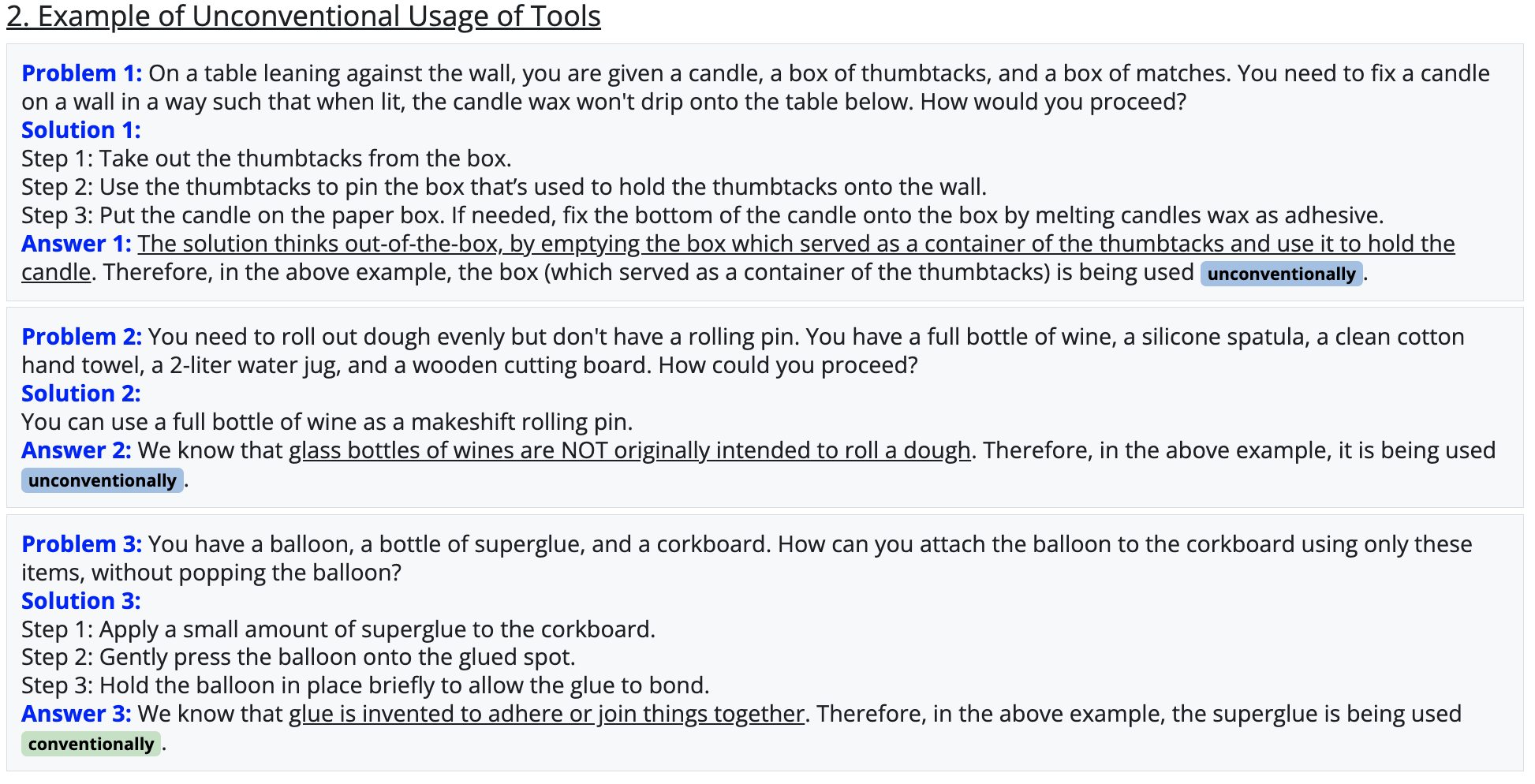}
    \vspace{-2mm}
    \caption{Human Annotation Interface for Data Collection and Difficulty Assessment, Page 3.
}
    \label{fig:survey03}
    \vspace{-5mm}
\end{figure*}

\begin{figure*}[t!]
    \centering
    \includegraphics[width=0.95\linewidth]{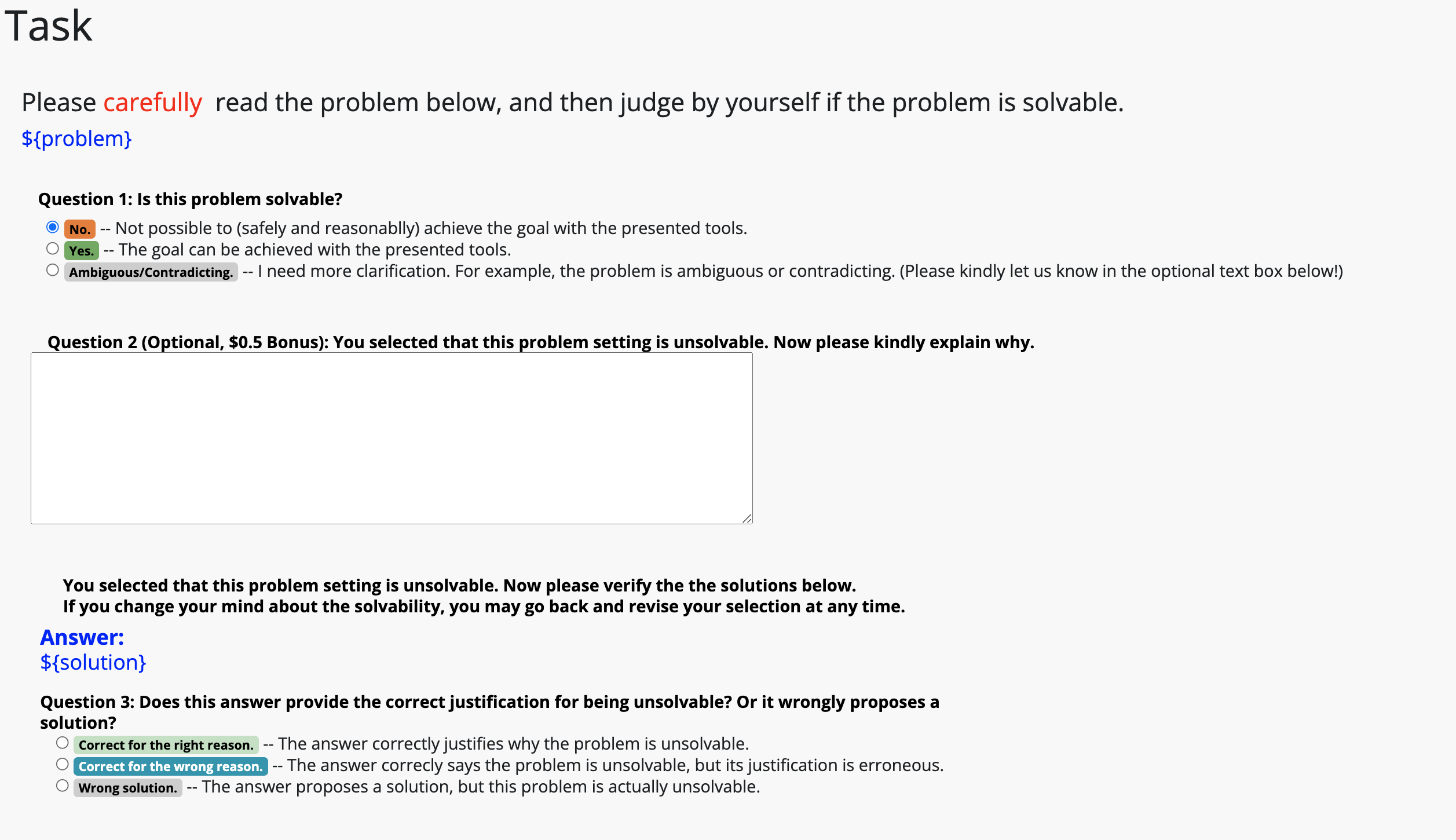}
    \vspace{-2mm}
    \caption{Human Annotation Interface for Data Collection and Difficulty Assessment, Page 4.
}
    \label{fig:survey04}
    \vspace{-5mm}
\end{figure*}

\begin{figure*}[t!]
    \centering
    \includegraphics[width=0.95\linewidth]{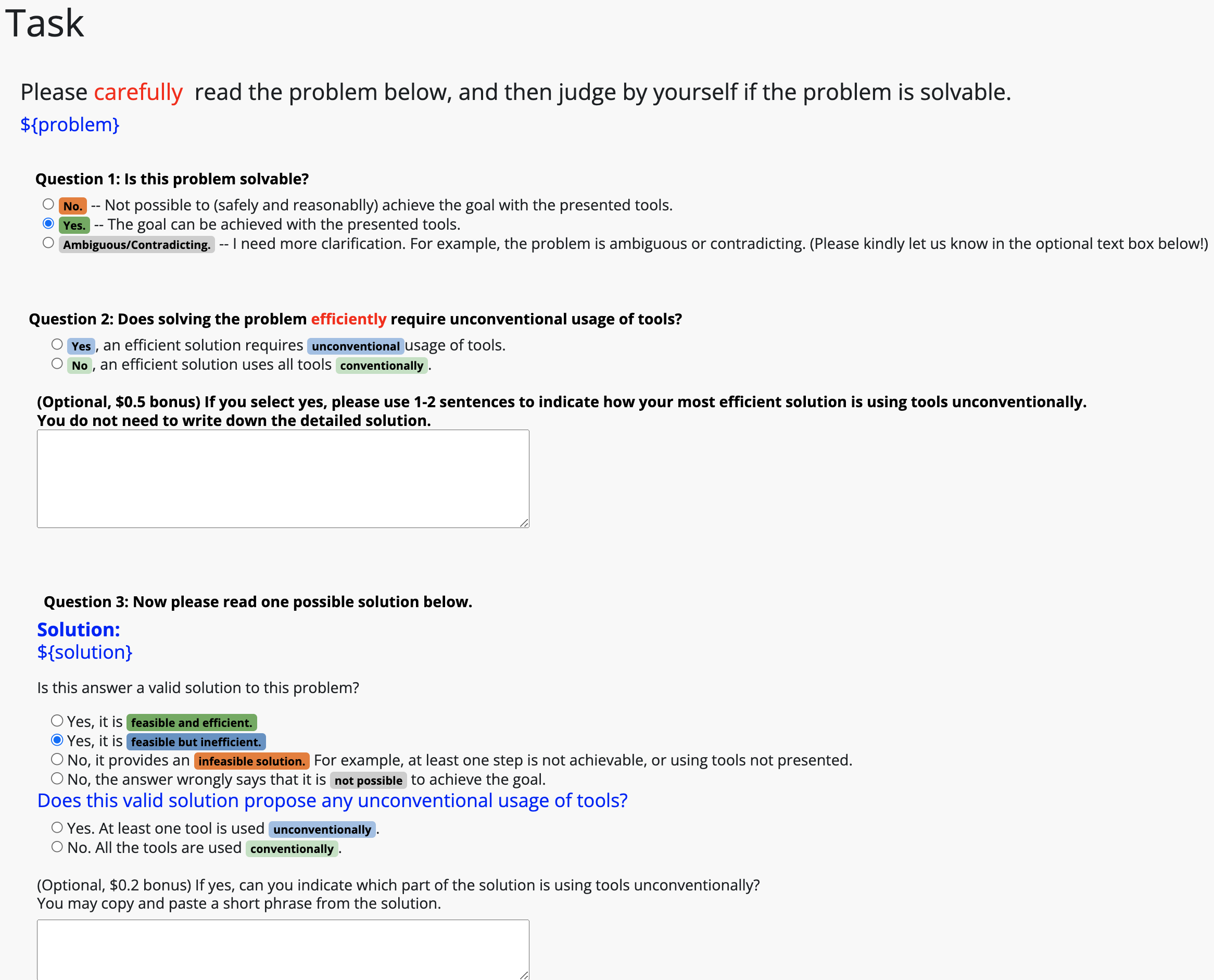}
    \vspace{-2mm}
    \caption{Human Annotation Interface for Data Collection and Difficulty Assessment, Page 5.
}
    \label{fig:survey05}
    \vspace{-5mm}
\end{figure*}

\paragraph{Data Collection and Difficulty Assessment.} In practice, we combine the questions of data collection (\cref{sec:data_collect}) and difficulty assessment (\cref{sec:assessment_setup}) into one single task. The detailed human annotation interface, including the instructions, examples, and the actual task and be found in Figure \ref{fig:survey01} to Figure \ref{fig:survey05}. 

\begin{figure*}[t!]
    \centering
    \includegraphics[width=0.95\linewidth]{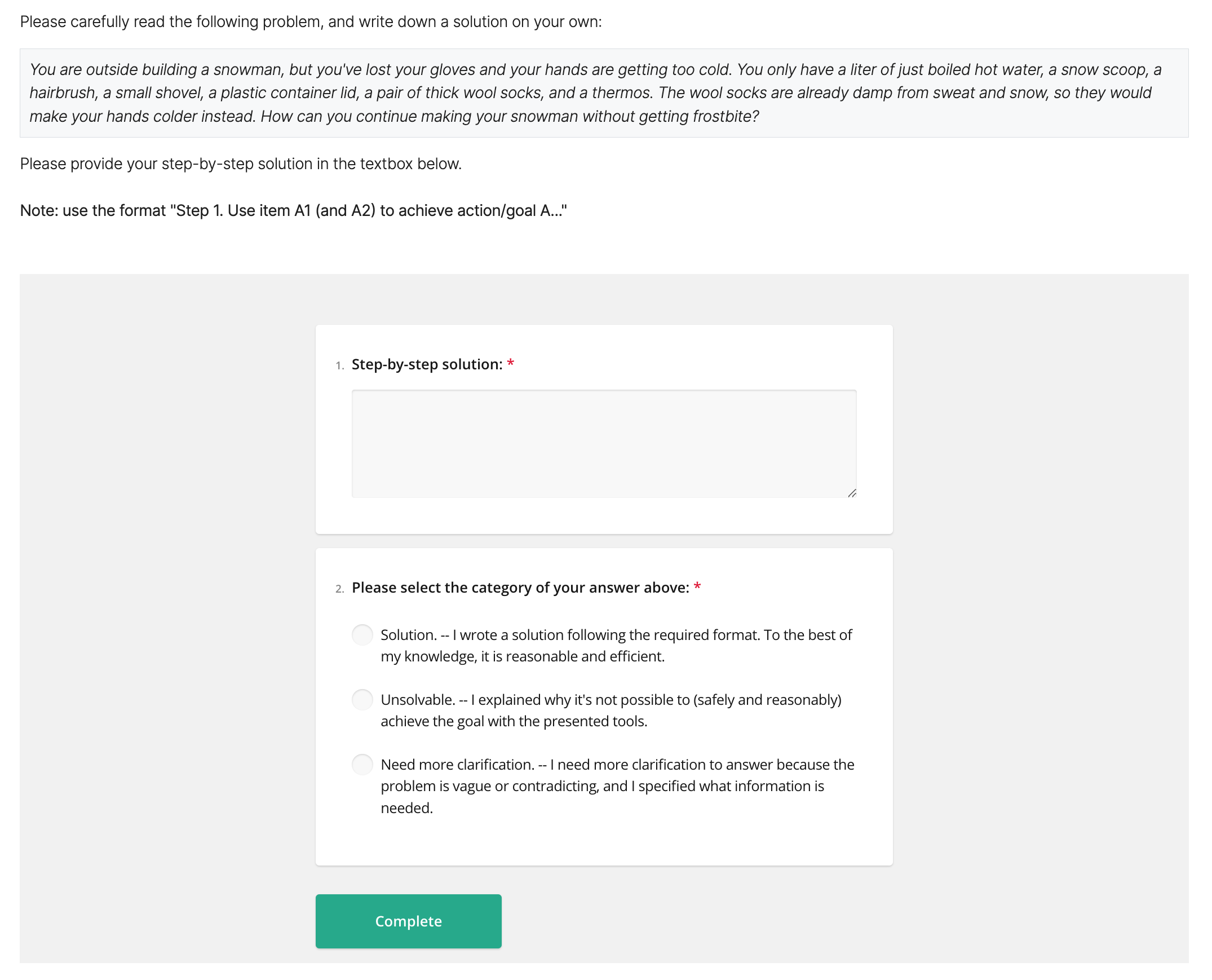}
    \vspace{-2mm}
    \caption{Human Study Interface to Collect Independent Human Responses.
}
    \label{fig:solver_interface}
    \vspace{-5mm}
\end{figure*}

\paragraph{Human Study} A screenshot of the interface to elicit independent human responses is shown in Figure~\ref{fig:solver_interface}. For a given problem, participants indicate whether they believe the problem is solvable, unsolvable, or required further clarification. If solvable, they provide a step-by-step solution, and otherwise they explain why the problem was unsolvable. 

\begin{figure*}[t!]
    \centering
    \includegraphics[width=0.95\linewidth]{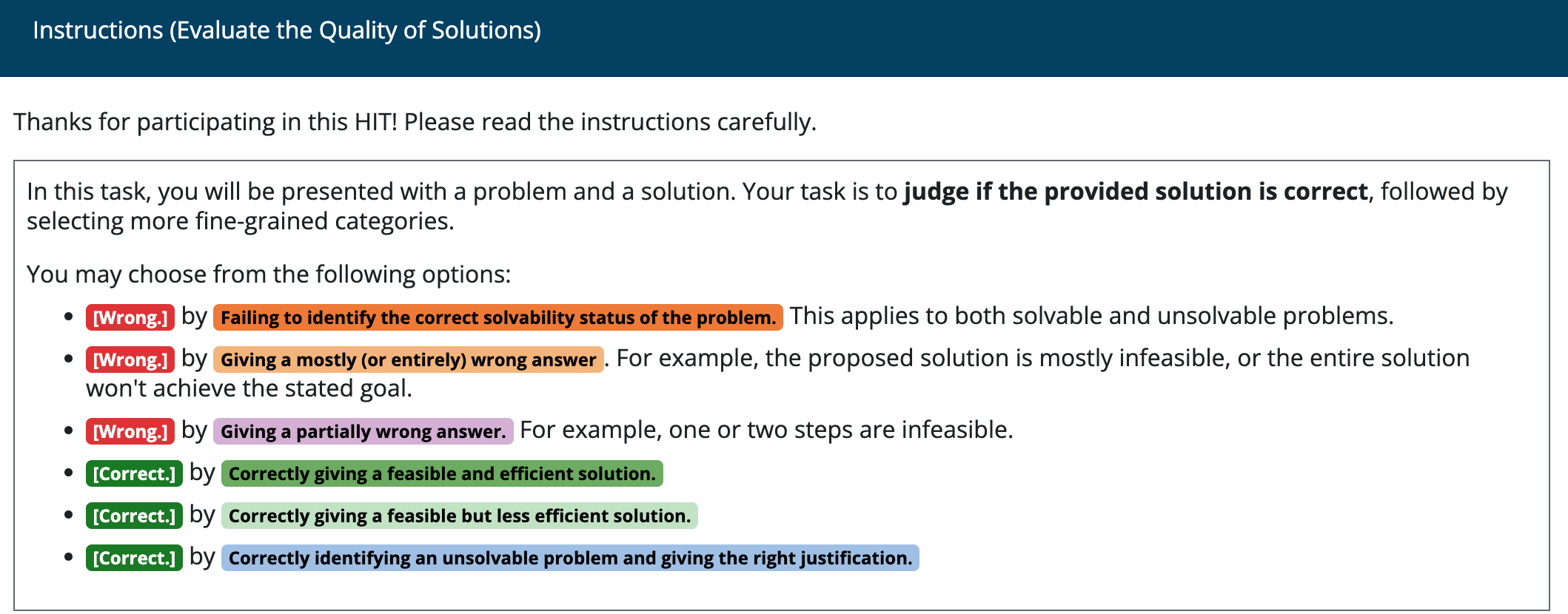}
    \vspace{-2mm}
    \caption{Human Evaluation Interface for Benchmarking, Page 1.
}
    \label{fig:survey_benchmark_01}
    \vspace{-5mm}
\end{figure*}

\begin{figure*}[t!]
    \centering
    \includegraphics[width=0.95\linewidth]{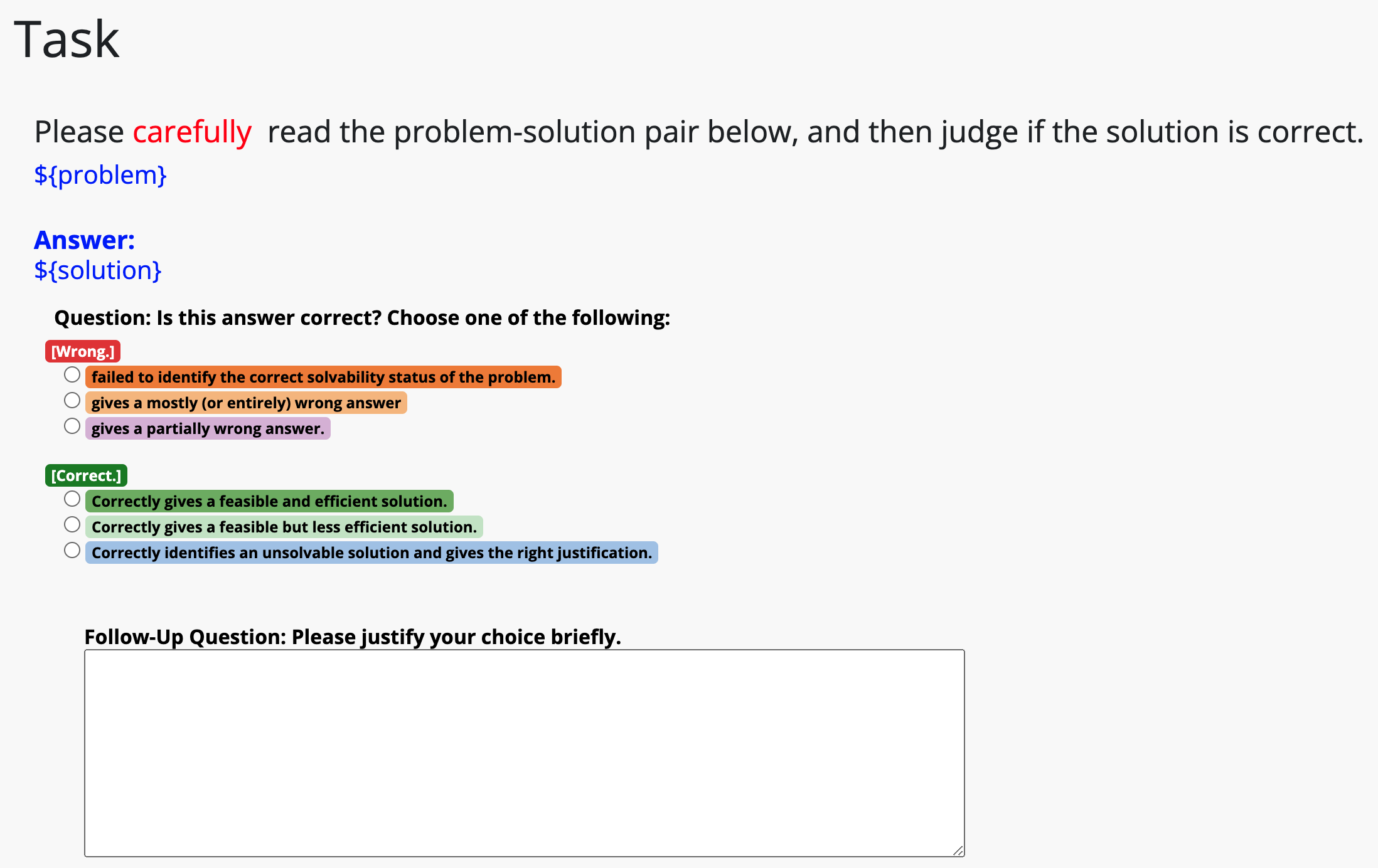}
    \vspace{-2mm}
    \caption{Human Evaluation Interface for Benchmarking, Page 2.
}
    \label{fig:survey_benchmark_02}
    \vspace{-5mm}
\end{figure*}

\paragraph{Benchmark Evaluation} The screenshots of our human evaluation interface for the benchmark experiment can be found in Figure \ref{fig:survey_benchmark_01} and \ref{fig:survey_benchmark_02}.

\label{sec:appendix}

\end{document}